\documentclass[10pt,twocolumn,letterpaper]{article}

\usepackage{cvpr}      
\usepackage[accsupp]{axessibility}

\usepackage[ruled,vlined]{algorithm2e}
\definecolor{cvprblue}{rgb}{0.21,0.49,0.74}
\definecolor{myred}{HTML}{880000}
\definecolor{mygreen}{HTML}{008800}
\definecolor{mygrey}{HTML}{656565}
\usepackage{times}
\usepackage{epsfig}
\usepackage{graphicx}
\usepackage{amsmath}
\usepackage{amssymb}
\usepackage{bm}
\usepackage{booktabs}
\usepackage{pifont}
\usepackage[pagebackref,breaklinks,colorlinks,allcolors=cvprblue]{hyperref}
\usepackage{makecell}
\usepackage{multirow}
\usepackage{tabularx,array}
\usepackage[table,xcdraw]{xcolor}
\usepackage{cvpr}              
\usepackage{times}
\usepackage{epsfig}
\usepackage{graphicx}
\usepackage{amsmath}
\usepackage{amssymb}
\usepackage{booktabs}
\usepackage{multirow}
\usepackage{enumitem}
\usepackage{mathtools}
\usepackage{bm}
\usepackage{bbm}
\usepackage{nicefrac}
\usepackage{array}
\usepackage{makecell}
\usepackage{caption}
\usepackage{subcaption}
\usepackage{siunitx}

\usepackage{algorithm}
\usepackage{algorithmic}
\usepackage{xcolor}
\usepackage{url}
\usepackage{tikz}
\usepackage{tikz}
\usepackage{graphicx} 
\usetikzlibrary{arrows.meta}
\usetikzlibrary{positioning,arrows.meta,shapes.multipart}
\usepackage{cuted}

\newcolumntype{C}{>{\centering\arraybackslash}X}
\def\paperID{} 
\def\confName{CVPR}
\def\confYear{2026}

\title{ReMoGen: Real-time Human Interaction-to-Reaction Generation via Modular Learning from Diverse Data}

\author{
    Yaoqin Ye$^{1}$,
    Yiteng Xu$^{1}$,
    Qin Sun$^{2,3}$,
    Xinge Zhu$^{4}$,
    Yujing Sun$^{5}$,
    Yuexin Ma$^{1,\dagger}$
    \\
    $^{1}$ShanghaiTech University,
    $^{2}$Guangzhou Institute of Energy Conversion, CAS,\\
    $^{3}$University of Science and Technology of China,
    $^{4}$The Chinese University of Hong Kong,\\
    $^{5}$Nanyang Technological University\\
    {\tt\small \{yeyq2025, mayuexin\}@shanghaitech.edu.cn}}

\begin{document}

\makeatletter
\let\@oldmaketitle\@maketitle
\renewcommand{\@maketitle}{
   \@oldmaketitle
 \begin{center}
   \vspace{-5ex}
      \includegraphics[width=1.0\linewidth]{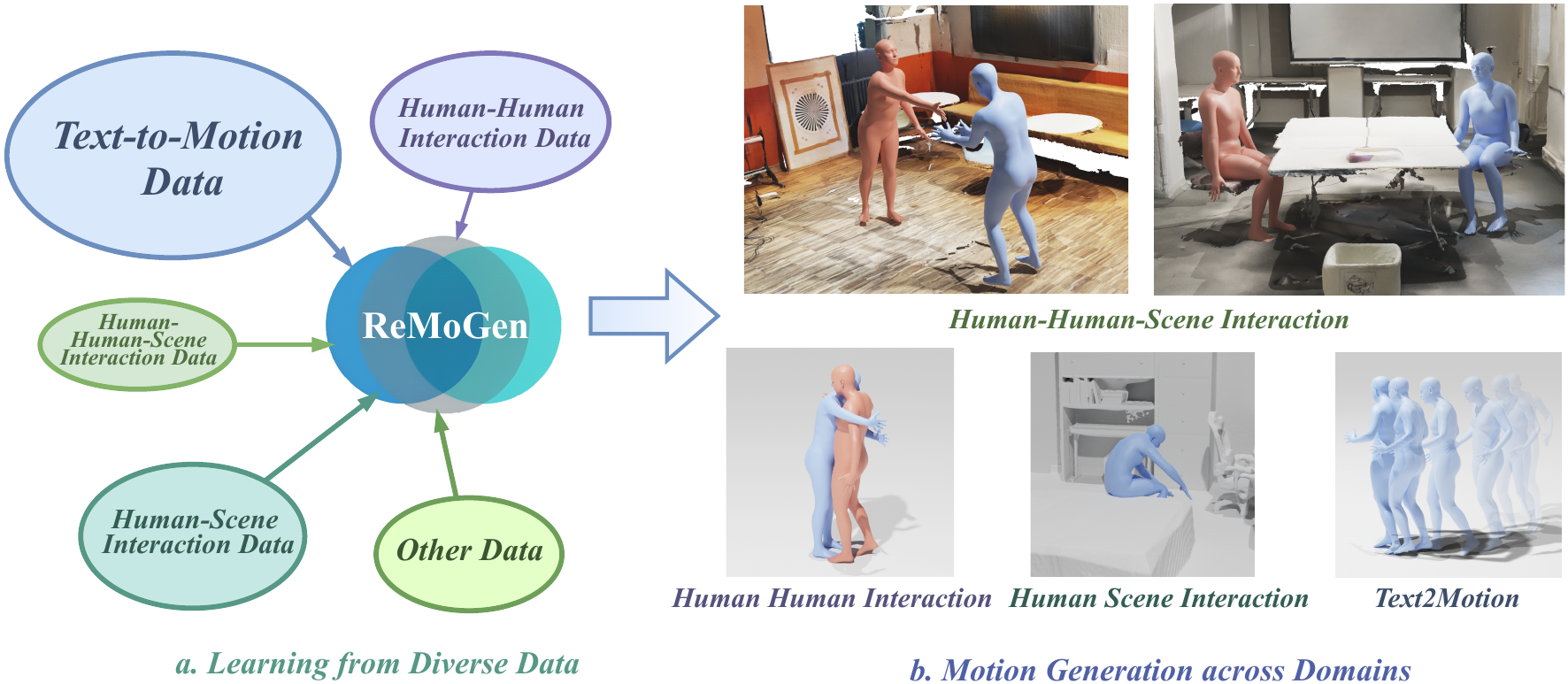}
 \end{center}
 \vspace{-2ex}
  \refstepcounter{figure}\normalfont Figure~\thefigure. 
  \textbf{Overview of ReMoGen.} ReMoGen is a modular framework that learns from heterogeneous interaction data. It supports real-time, high-quality, and coherent reaction generation across both single-domain and mixed-modality interaction settings. Our project page and code can be found at \url{https://4dvlab.github.io/project_page/remogen/}.
  \label{fig:teaser}
  \newline
  }
\maketitle
\footnotetext[1]{$\dagger$ Corresponding author.}
\begin{abstract}
Human behaviors in real-world environments are inherently interactive, with an individual's motion shaped by surrounding agents and the scene. Such capabilities are essential for applications in virtual avatars, interactive animation, and human-robot collaboration. 
We target \textbf{real-time human interaction-to-reaction generation}, which generates the ego's future motion from dynamic multi-source cues, including others' actions, scene geometry, and optional high-level semantic inputs.
This task is fundamentally challenging due to (i) limited and fragmented interaction data distributed across heterogeneous single-person, human-human, and human-scene domains, and (ii) the need to produce low-latency yet high-fidelity motion responses during continuous online interaction. 
To address these challenges, we propose \textbf{ReMoGen} (\textbf{Re}action \textbf{Mo}tion \textbf{Gen}eration), a modular learning framework for real-time interaction-to-reaction generation. 
ReMoGen leverages a universal motion prior learned from large-scale single-person motion datasets and adapts it to target interaction domains through independently trained \textbf{Meta-Interaction} modules, enabling robust generalization under data-scarce and heterogeneous supervision.
To support responsive online interaction, ReMoGen performs segment-level generation together with a lightweight \textbf{Frame-wise Segment Refinement} module that incorporates newly observed cues at the frame level, improving both responsiveness and temporal coherence without expensive full-sequence inference.
Extensive experiments across human-human, human-scene, and mixed-modality interaction settings show that ReMoGen produces high-quality, coherent, and responsive reactions, while generalizing effectively across diverse interaction scenarios. 

\end{abstract}    
\section{Introduction}
\label{sec:intro}
Human behaviors in realistic environments are inherently interactive and context-dependent~\cite{Xu2024ReGenNet, Chen2024SymBridge}. 
An individual's action is often a \emph{reaction}—dynamically shaped by the observed motions of other people and the surrounding scene.
We target the task of \textbf{real-time human interaction-to-reaction generation} as predicting the ego's future motion conditioned on multi-source interaction cues, including the behaviors of surrounding agents, the spatial configuration of the scene, and high-level textual cues that specify interaction intent.
This task requires the model to continuously perceive and respond to dynamic human-human-scene interactions, generating physically plausible and semantically consistent motion in real time. Accurately modeling such interaction-to-reaction dynamics is a key step toward embodied intelligence that can understand and respond to complex social and physical contexts. 
It enables lifelike digital avatars and cooperative robots to exhibit human-level reactivity and situational awareness. 
Applications span from \emph{virtual character animation, gaming, and film production}~\cite{Tevet2023MDM, Zhao2024DART} to \emph{assistive robotics, human-robot collaboration}~\cite{liu2025core4d,Chen2024SymBridge}, and other interactive AI systems. 

Human motion generation has evolved from text-driven single-person actions~\cite{Guo2022HumanML3D, Petrovich2022TEMOS} to context-aware and multi-agent scenarios~\cite{Wang2023InterControl, Jiang2024TRUMANS,Fan2024FreeMotion}. 
\textit{Text-to-motion} methods learn direct mappings from natural language to human motion, producing diverse and semantically aligned sequences~\cite{Guo2022HumanML3D, Petrovich2022TEMOS, Tevet2023MDM} but in isolation from the physical world. 
\textit{Human-scene interaction} approaches~\cite{Jiang2024TRUMANS,Jiang2024Autonomous,Milacski2024GHOST} introduce spatial awareness by conditioning on 3D layouts or scene geometry, yet remain limited to single-agent behaviors. 
\textit{Human-human interaction} frameworks~\cite{Liang2023InterGen, Wang2023InterControl, Fan2024FreeMotion} attempt to generate all interacting agents jointly under pre-defined interaction constraints, while \textit{action-to-reaction} settings~\cite{Xu2024ReGenNet,Chen2024SymBridge} predict how one agent reacts to another. Despite these advances, these paradigms have not captured the essence of \textbf{real-time interaction-to-reaction generation}, where an agent has to respond instantly to dynamic inputs from others and the environment. Most of these still assume paired or fully observed motion sequences and operate \textit{offline}~\cite{Fan2024FreeMotion,Chen2025Sitcom}, observing both past and future actions of others before generation. 

Although our target task captures a natural aspect of human behavior, it introduces two fundamental challenges that current motion-generation frameworks fail to address.

\noindent\textit{(1) Data scarcity and heterogeneity.}
Available motion datasets vary widely in domain coverage and supervision modality—ranging from large-scale single-person actions~\cite{Guo2022HumanML3D, punnakkal2021babel} to smaller and more fragmented human-human~\cite{fieraru2020three,Liang2023InterGen, Xu2024InterX} and human-scene interaction sets~\cite{cong2024laserhuman,Jiang2024TRUMANS,Jiang2024Autonomous}.  
This combination of limited interaction data and strong heterogeneity makes it difficult to learn unified motion priors or transferable representations.  
As a result, end-to-end models trained within a single domain often overfit to its specific distribution and generalize poorly to new interaction configurations.  
Effectively exploiting all available yet sparse and domain-specific resources remains an open problem for building a generalizable interaction-to-reaction model.

\noindent
\textit{(2) Real-time responsiveness.}
Practical applications such as virtual agents, interactive avatars, embodied assistants, and human-robot collaboration~\cite{Chen2024SymBridge} require motion responses that are both \emph{immediate} and \emph{high-quality}.  
However, existing architectures struggle to maintain this balance.  
Diffusion-based models~\cite{Tevet2023MDM,Chen2023MLD, Yuan2023PhysDiff} achieve strong motion fidelity but depend on offline generation or heavy optimization, making them too slow for interactive use.  
Conversely, online autoregressive designs~\cite{Zhao2024DART, Chen2024CAMDM} offer higher speed but often accumulate prediction errors, leading to degraded realism and noticeable drift over long horizons.  
Achieving \textbf{low-latency reactions without sacrificing motion fidelity} remains a central technical challenge, as models have to continuously integrate dynamic observations while preserving smooth, coherent, and physically plausible motion.

To address the above challenges, we propose \textbf{ReMoGen} (\textbf{Re}action \textbf{Mo}tion \textbf{Gen}eration), 
a \textbf{modular learning framework} for real-time human interaction-to-reaction generation. 
ReMoGen addresses reaction generation through a modular design that separates universal motion prior learning from interaction-specific adaptation. It supports diverse interaction settings by combining a shared motion prior with domain-specific modules that incorporate observations of other agents and the scene. Specifically, ReMoGen consists of two functional layers that address the two major challenges:

\begin{enumerate}
    \item {\textbf{Prior-guided Modular Learning for Heterogeneous Interaction Domains.}
    
ReMoGen learns a universal motion prior from abundant single-person datasets and transfers it to scarce interaction domains through independently trained \emph{Meta-Interaction} modules.  
Rather than relying on extensive joint multi-domain training, these modules rapidly adapt to new target domains by leveraging the shared motion prior.
In this way, ReMoGen transfers general motion knowledge learned from single-person data to interaction settings, and further generalizes to more complex or previously unseen scenarios, effectively addressing the challenges of data heterogeneity and data scarcity.}

\item {\textbf{Frame-wise Segment Refinement for Real-time Rollout.} 

During rollout, ReMoGen performs real-time generation in short temporal segments instead of processing full sequences through the large backbone network.  
Within each segment, a lightweight \emph{Frame-wise Segment Refinement} module updates motion predictions using the latest incoming observations of other agents and the scene.  
This refinement mechanism provides a balance between motion fidelity and low-latency responsiveness: 
the large backbone offers stable long-term dynamics, while the small segment-wise adapter ensures rapid, fine-grained reactivity.
}
\end{enumerate}
In summary, our main contribution is a unified framework for \textbf{real-time human interaction-to-reaction generation} that addresses data scarcity, heterogeneous supervision, and real-time responsiveness in a modular design. \textbf{ReMoGen} achieves this by combining a universal motion prior with independently trained \emph{Meta-Interaction Modules} for efficient interaction-specific adaptation, together with a lightweight \emph{Frame-wise Segment Refinement} mechanism for responsive online generation. Extensive experiments show that ReMoGen produces high-quality, coherent, and responsive reactions across diverse interaction scenarios and input modalities, establishing an effective and scalable approach to real-time interaction-to-reaction generation.

\begin{figure*}[t]
    \centering
      \includegraphics[width=\linewidth]{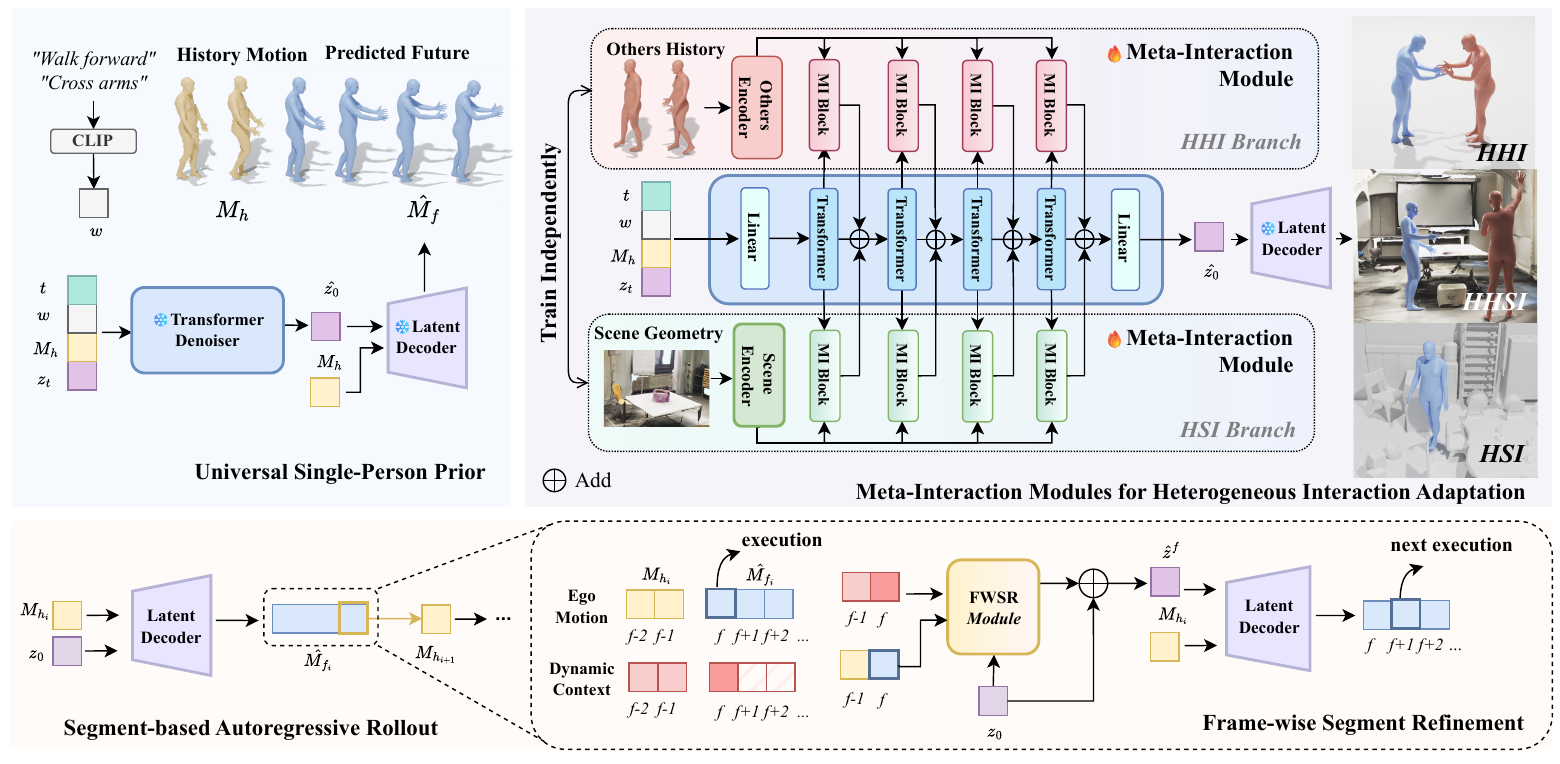}
       \caption{\textbf{Overview of the ReMoGen Framework}.
    Our framework is designed to address the challenges of data scarcity and real-time responsiveness in interaction-to-reaction generation.
     }
    \label{fig:mainfig}
\end{figure*}

\section{Related Work}

\subsection{Single-Person Text-to-Motion Generation}

Single-person text-to-motion generation has progressed rapidly.  
Early VAE-based models such as T2M~\cite{Guo2022HumanML3D} and TEMOS~\cite{Petrovich2022TEMOS} demonstrated effective language-to-motion mapping~\cite{hassan2021stochastic, petrovich2021action}, and diffusion-based approaches~\cite{Tevet2023MDM, kim2022flame, zhang2022motiondiffuse, dabral2023mofusion} further improved realism and diversity.  
Subsequent developments explored latent diffusion for efficiency~\cite{Chen2023MLD, ren2023diffusion}, incorporated physical constraints for more plausible synthesis~\cite{Yuan2023PhysDiff, Tevet2024CLOSD}, and introduced lightweight autoregressive architectures for real-time generation~\cite{Zhao2024DART, Chen2024CAMDM, shi2024interactive}.

Despite this progress, single-person models largely operate in isolation from social context.  
They produce semantically accurate motion but lack awareness of surrounding agents or scene geometry, limiting their use in in-teractive settings.  
Nevertheless, large-scale single-person datasets~\cite{Guo2022HumanML3D, punnakkal2021babel, mahmood2019amass, plappert2016kit} and the models trained on them provide rich motion priors capturing kinematic structure, temporal regularity, and semantic diversity, offering strong potential for transfer to downstream interaction tasks.

\subsection{Human Interaction Motion Generation}
Human interaction motion generation extends the single-person text-to-motion task to scenarios involving multiple agents and scene context. This task has progressed across Human-Scene Interaction (HSI), Human-Human Interaction (HHI), and broader multi-entity settings.

HSI methods model how humans navigate or interact with 3D environments using scene geometry or affordances~\cite{Jiang2024TRUMANS,Jiang2024Autonomous, Milacski2024GHOST, Wang2024Move, hassan2023synthesizing, zhao2023synthesizing}.
HHI approaches focus on interpersonal coordination, ranging from collaborative behaviors to fine-grained action-reaction dynamics~\cite{Liang2023InterGen, Xu2024InterX, Wang2023InterControl, Xu2024ReGenNet, Ghosh2024REMOS, Fan2024FreeMotion, barquero2024seamless, shan2024towards}.
Recent actor-reactor frameworks generate reactions in response to another's motion~\cite{Xu2024ReGenNet, Fan2024FreeMotion,liu2025uni}, while real-time online systems demonstrate the feasibility of producing responsive interactions in human-robot collaboration~\cite{Chen2024SymBridge}. 

Despite these developments, many frameworks rely on domain-specific datasets, treat HHI and HSI separately, or require fully observed partner trajectories~\cite{guo2020action2motion, li2021ai}.
Recently, a notable attempt to address human-human-scene interaction~\cite{Chen2025Sitcom} assembles separate specialized modules to produce in-scene multi-person motion.
While it brings interaction into scene context, it operates strictly offline and merges independently generated segments via interpolation, limiting adaptability to dynamic interaction cues.
 As a result, real-time interaction-to-reaction generation that remains responsive, scene-aware, and stable under evolving conditions is still insufficiently addressed.

\subsection{Prior-Guided Adaptation in Motion Generation}

A growing direction in generative modeling adapts large pre-trained backbones using lightweight modular components.
In 2D image synthesis, approaches such as LoRA~\cite{hu2022lora}, ControlNet~\cite{Zhang2023Adding}, and T2I-Adapter~\cite{Mou2023T2IAdapter} demonstrate that attaching small trainable branches or parameter-efficient layers to a frozen prior enables flexible conditioning without retraining the full model.
This philosophy—leveraging strong foundational priors while introducing task-specific adapters—has begun influencing 3D motion generation.

Modular motion approaches adopt this idea in several ways.  
Some introduce spatial or geometric control branches that guide the backbone with scene or contact cues~\cite{Zhao2024DART, Wang2023InterControl, karunratanakul2024optimizing}.  
Others employ noise-based optimization modules that modify diffusion noise or latent trajectories to impose spatial or interaction constraints at runtime~\cite{karunratanakul2024optimizing, wallace2023end}.  
Another line of work decomposes motion generation into separate functional components or hierarchical body parts, offering improved controllability and flexibility~\cite{Ghosh2024REMOS, Fan2024FreeMotion, Tang2024PointPEFT, zhang2023finemogen}.

While these strategies highlight the promise of prior-guided modularity, they are typically domain-specific, operate offline, or rely on fully observed partner trajectories, and therefore do not directly address how to reuse a shared motion prior across heterogeneous interaction domains under sparse supervision.  
In contrast, ReMoGen is designed to transfer general motion knowledge learned from single-person data to interaction domains through independently trained modules, and further support responsive online generation through frame-wise refinement.  
This distinction makes ReMoGen suitable for real-time interaction-to-reaction generation across diverse interaction scenarios and input modalities.

\section{Method}
\subsection{Overview}

ReMoGen is a modular framework for real-time interaction-to-reaction generation (Fig.~\ref{fig:mainfig}). Given text intent, observed motions of other agents, and scene context, it autoregressively predicts the ego's future motion using three components: (1) a frozen text-conditioned single-person motion prior pretrained on large-scale datasets; (2) Meta-Interaction modules that adapt this prior to different interaction domains (e.g., HHI, HSI); and (3) a Frame-wise Segment Refinement module that updates short predicted segments using the latest interaction cues to achieve low-latency online rollout while preserving motion fidelity.

\begin{figure*}[t]
  \centering
  \includegraphics[width=\linewidth]{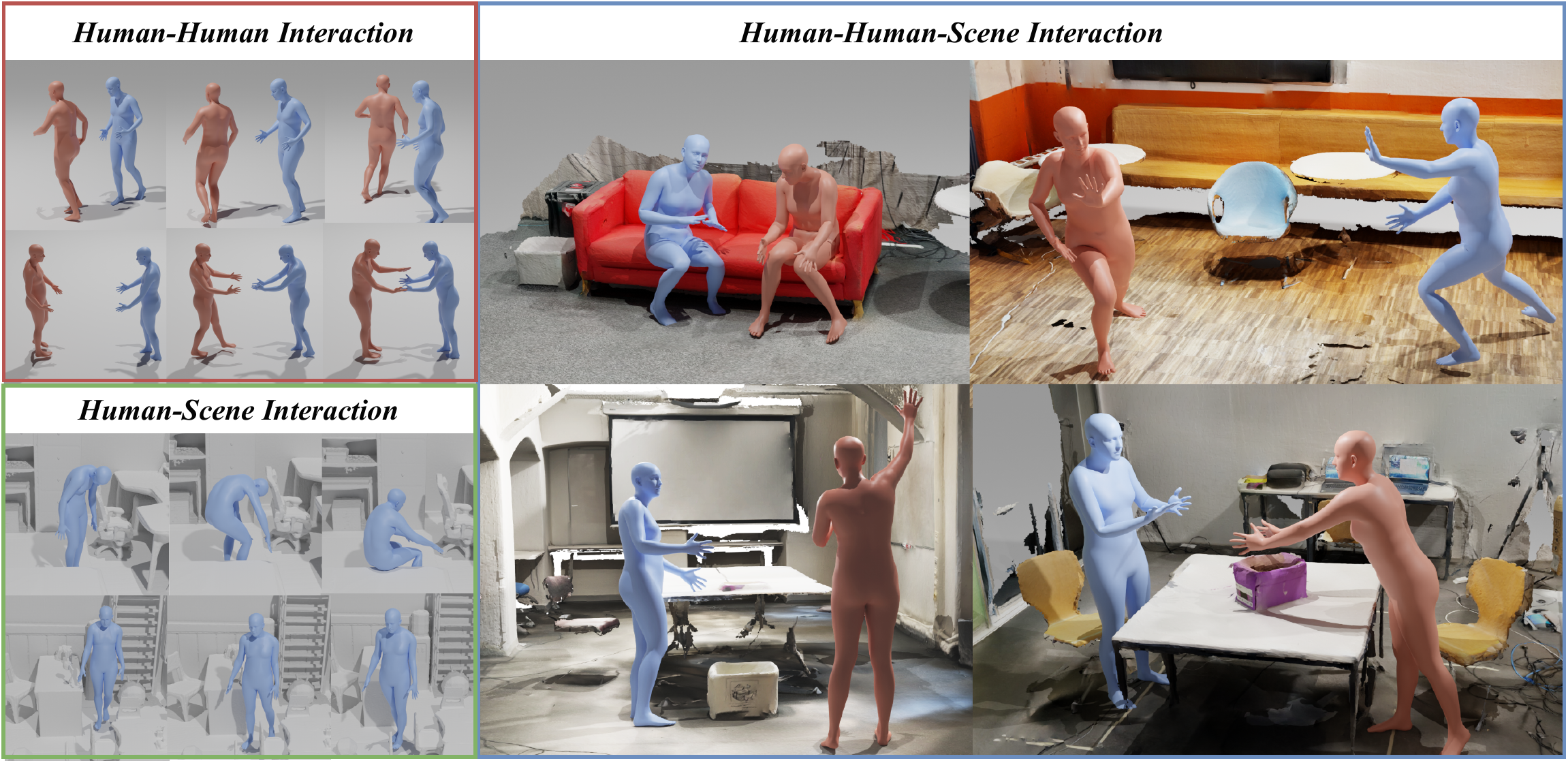}
       \caption{Qualitative results of ReMoGen across Human--Human, Human--Scene, and mixed Human--Human--Scene scenarios. Blue meshes denote the generated ego motion, while red meshes represent the observed motions of others. The examples cover diverse interaction behaviors, including Taichi-style movements, chasing, chatting, and scene-aware interaction, demonstrating the versatility of ReMoGen across heterogeneous interaction settings.}
   \label{fig:vis}
\end{figure*}
\subsection{Universal Single-Person Prior}

ReMoGen builds upon a universal text-conditioned single-person motion prior learned from large-scale text-to-motion datasets. This prior captures general kinematic structure, temporal dynamics, and language-motion correlations. It is kept frozen during all interaction-aware learning, serving as the backbone motion generator for ReMoGen.

\textbf{Problem Formulation.} Given a textual description \(W\), the prior generates a sequence of 3D human poses \(M = \{m^{(1)}, \dots, m^{(T)}\}\), where each pose \(m^{(i)} \in \mathbb{R}^D\) follows a modified SMPL-X~\cite{pavlakos2019expressive} representation encoding body pose, hand pose, root motion, and other dynamic attributes~\cite{Zhao2024DART}.

\textbf{Segment-based Autoregressive Generation.} For efficient long sequences, we adopt a segment-based autoregressive approach: the model iteratively predicts a future segment \(\hat{M}^i_f \in \mathbb{R}^{F \times D}\) from a history window \(M^i_h \in \mathbb{R}^{H \times D}\) and text \(W\), and updates the history via
\[
M^{i+1}_h = \mathrm{concat}(M^i_h, \hat{M}^i_f)[-H:].
\tag{1}
\]

\textbf{Latent Motion Diffusion.} Following DART~\cite{Zhao2024DART}, each segment is encoded into a latent space with a transformer-based VAE:
\[
\hat{M}^i_f = D_\theta(M^i_h, z^i),
\tag{2}
\]
where \(w = \mathrm{Enc}_{\text{text}}(W)\) conditions a latent denoiser \(G_\psi\). At diffusion step \(t\), the denoiser predicts a clean latent
\[
\hat{z}_0 = G_\psi(z_t, t, M^i_h, w),
\tag{3}
\]
which is decoded back to motion space by the VAE decoder.

\subsection{Meta-Interaction Modules for Heterogeneous Interaction Adaptation}
To imbue the universal prior with interaction awareness, we introduce \textbf{Meta-Interaction Modules}. These adapters plug into the frozen prior and inject guidance from dynamic interaction cues without requiring expensive full-model retraining. While high-level semantic inputs are consumed only by the universal prior, Meta-Interaction Modules remain text-agnostic and specialize in adapting the shared prior to a particular interaction domain (e.g., HHI, HSI) based on observations of other agents and the scene. Each module is trained independently.

\paragraph{Overview.}
At each denoising step \(t\), the latent denoiser receives: (i) the ego motion history $M_h^{i}$, (ii) a noisy latent $z_t$, and (iii) contextual embeddings derived from either other agents or the scene.  
Meta-Interaction Modules transform these contextual embeddings into adaptive modulation signals that steer the prior's latent update:
\[
\hat{z}_0 = 
G_\psi\!\left(z_t,\ t,\ M_h^{i},\ 
c_{\text{others}},\ 
c_{\text{scene}},\ 
w\right)
\tag{4}
\]
where text embedding $w$ conditions the \textbf{universal prior only}, while $c_{\text{others}}$ and $c_{\text{scene}}$ exclusively drive the Meta-Interaction Modules.

\paragraph{Context Encoding.}
Two independent encoders extract high-level interaction cues:
\[
c_{\text{context}} = \mathrm{Enc}_{\text{context}}(X_{\text{context}})
\tag{5}
\]
where $X_{\text{others}}$ is a short history of surrounding agents' poses, $X_{\text{scene}}$ is a voxelized 3D occupancy representation of the environment.

The \textit{Others Encoder} (TCN-based)~\cite{Chen2024SymBridge} captures relative velocity, approach direction, and human-human spatial relations, while the \textit{Scene Encoder} (ViT-based)~\cite{Jiang2024Autonomous} summarizes nearby geometry and affordances. Both encoders are modular and replaceable.

\subsection{Meta-Interaction Block}

The Meta-Interaction Block injects interaction-awareness into the frozen prior. Given ego features \(h\) and contextual cues \(c_{\text{context}}\), a self-attention layer first refines \(h\) into \(h'\), and a cross-attention layer over \(c_{\text{context}}\) extracts interaction-relevant signals
. These signals are converted into FiLM-style modulation parameters \((\gamma, \beta)\) that apply a feature-wise affine transformation
\[
h_{\text{mod}} = (1 + \tanh \gamma) \odot h' + \tanh \beta,
\tag{6}
\]
followed by a feed-forward layer to obtain interaction-aware features for each denoising step.

\begin{algorithm}[b]
\caption{Frame-wise Segment Refinement}
\label{alg:framewise}
\KwIn{
latent $z_0$,  
dynamic interaction cues $X_{\text{dyn}}$, 
initial future segment $\hat{M}_f^{(0)}$,
ego history window $M_h$, 
decoder $D_\theta$, 
frame-wise adapter \textsc{FWSR}.
}
\KwOut{Refined future segment $\hat{M}_f$.}

\BlankLine
Initialize first-frame prediction: $m^{(0)} \leftarrow \hat{M}_f^{(0)}[:,1,:]$ \\
Update history window:
$M_h^{(0)} \leftarrow \text{concat}(M_h[:,2:H,:],\, m^{(0)})$ \\

\For{$f = 1$ \KwTo $F-1$}{
    Extract recent dynamic window: \\
    $X_{\text{dyn}}^{(f)} \leftarrow X_{\text{dyn}}[:, -H:, :]$ \\

    Refine latent:
    $\tilde{z}^f \leftarrow \textsc{FWSR}(z_0,\, M_h^{(f-1)},\, X_{\text{dyn}}^{(f)})$ \\

    Decode refined future segment:
    $\tilde{M}_f \leftarrow D_\theta(M_h^{(0)},\, \tilde{z}^f)$ \\

    Take next frame:
    $m^{(f)} \leftarrow \tilde{M}_f[:, 1+f, :]$ \\

    Update history:
    $M_h^{(f)} \leftarrow \text{concat}(M_h^{(f-1)}[:,2:H,:],\, m^{(f)})$ \\
}
\Return $\hat{M}_f = \text{concat}(m^{(0)}, m^{(1)}, \ldots, m^{(F-1)})$
\end{algorithm}

\paragraph{Training Strategy.}
Each Meta-Interaction Module is trained \textbf{independently} on its corresponding domain (HHI on Inter-X, HSI on LINGO) while the universal prior remains frozen. This allows each module to specialize without requiring costly joint supervision. The training objectives mirror those of the prior~\cite{Zhao2024DART}, including reconstruction and temporal-delta losses (\(L_{\text{rec}}, L_{\text{latent}}, L_{\text{aux}}\)), ensuring that the learned modulations are domain-consistent and temporally stable.

\paragraph{Composable Inference.}
At inference time, each Meta-Interaction Module produces a modulation term \(\Delta_i\) that reflects the influence of its corresponding interaction cue on the ego latent representation.  
Multiple modules can be activated simultaneously, and their effects are combined through a weighted sum:

\[
\Delta_{\text{total}} = \sum_i \alpha_i \Delta_i,
\tag{7}
\]

where \(\alpha_i\) are user-defined coefficients that control the relative contribution of each domain.

To prevent instability from excessively large updates, the fused modulation \(\Delta_{\text{total}}\) is normalized with an L2-norm clamp before being applied to the latent features.  
This composable design enables flexible blending of HHI and HSI cues while preserving the integrity of the frozen prior.

\subsection{Frame-wise Segment Refinement for Low-Latency Online Updates}

Standard segment-based autoregression presents a trade-off: long segments ensure smooth, high-quality motion but introduce significant latency, as updates occur only after a full segment is generated. Conversely, short segments reduce latency but can lead to jerky, incoherent motion. Recomputing the entire diffusion process for every frame is computationally infeasible for real-time applications. The central challenge, therefore, is to bridge the gap between \emph{high-fidelity generation} and \emph{low-latency response}.

\paragraph{Frame-wise Segment Refinement (FWSR).}
To resolve this trade-off, we introduce Frame-wise Segment Refinement (FWSR), a lightweight module that provides per-frame corrective updates on top of the segment-level prediction. Instead of re-running the full generative process, FWSR efficiently adjusts the latent space. This is achieved using a lightweight version of our Meta-Interaction Block, applying a similar \texttt{Modulate} function with attention layers and FiLM. It uses the most recent dynamic context to refine the initial segment latent $z_0$:
\[
\hat{z}^f = 
\mathrm{Modulate}\big(z_0,\; \text{concat}(M_h^{(f-1)},\, X_{\mathrm{dyn}}^{(f)})\big),
\tag{8}
\]
where $z_0$ is the latent predicted by the main generator at the start of the segment. The dynamic context is formed by concatenating the most recent motion history $M_h^{(f-1)}$ and newly observed interaction cues $X_{\mathrm{dyn}}^{(f)}$. This refinement updates only the current frame, preserving long-horizon dynamics from the segment predictor while enabling immediate reaction to new observations.

FWSR is trained \textbf{independently} with the universal prior and all Meta-Interaction Modules frozen, ensuring it acts as a stable, local adapter that injects dynamic corrections without altering global motion structure. By iteratively refining frame-level predictions and updating the history buffer (Algorithm~\ref{alg:framewise}), FWSR achieves the responsiveness required for real-time interaction without sacrificing the quality of segment-level generation.
\begin{table*}[t]
\centering
\begin{tabularx}{\textwidth}{XCCCCCCC}
\toprule
Method            & FID $\downarrow$           & \makecell{R-Precision\\ \small{(Top-3)} $\uparrow$ }    & MM Dist.  $\downarrow$      & Diversity$\rightarrow$      & Peak Jerk$\downarrow$ & Collision(\%)$\downarrow$    & \makecell{Latency \\ \small{(s/frame)}$\downarrow$} \\ \midrule
\textcolor{mygrey}{GT}                & \textcolor{mygrey}{0.000}          & \textcolor{mygrey}{0.472}          & \textcolor{mygrey}{4.051}          & \textcolor{mygrey}{4.084}                   & \textcolor{mygrey}{0.343 }         & \textcolor{mygrey}{--}  & \textcolor{mygrey}{--}           \\\midrule
ReGenNet~\cite{Xu2024ReGenNet}          &       11.622         &       0.269         &       6.092         &           3.452           &       2.321       & 0.665 & \textcolor{myred}{0.210} \\
FreeMotion~\cite{Fan2024FreeMotion}& 3.383          & 0.284          & 5.438          & 3.394          & 2.207          & \textbf{0.545} &  \textcolor{myred}{0.221}            \\
\(\text{FreeMotion}^{\text{off}}\)~\cite{Fan2024FreeMotion}       & 0.492          & 0.417          & 4.330          & \underline{3.913}            & 0.304     &   1.220  &     --           \\
SymBridge~\cite{Chen2024SymBridge} & 2.569          & 0.355          & 4.955          & 3.598                 & \textbf{0.212} &  \underline{0.646}  &    \textcolor{mygreen}{0.040}        \\ \midrule
Ours              & \underline{0.181}          & \textbf{0.464} & \textbf{4.076} & 3.911        & 0.222    &   0.989   &   \textcolor{mygreen}{0.042}             \\
Ours+FWSR   & \textbf{0.166} & \underline{0.462}    & \textbf{4.076} & \textbf{4.003}      & \underline{0.218}  & 0.961 &    \textcolor{mygreen}{0.047}            \\ \bottomrule
\end{tabularx}
\caption{Human–Human Interaction results on Inter-X.
\textbf{Bold} indicates the best and \underline{underlined} indicates the second-best. 
Reported latency is end-to-end inference time. \textcolor{mygreen}{Green} latency values meet the 0.1\,s/frame real-time threshold (equivalent to 10 FPS), while \textcolor{myred}{red} values do not.}
    \label{tab:hhi}
\end{table*}
\section{Experiments}
We evaluate ReMoGen on real-time interaction-to-reaction generation across human-human (HHI), human-scene (HSI), and mixed human-human-scene settings. Our experiments are designed to answer four key questions:
\begin{itemize}
    \item \textbf{Q1:} Does our universal motion prior enable data-efficient adaptation to heterogeneous interaction domains?
    \item \textbf{Q2:} Do the Meta-Interaction Modules improve motion quality for conditioned interaction across HHI and HSI tasks?
    \item \textbf{Q3:} Can our composable design generalize to unseen mixed-modality interaction settings?
    \item \textbf{Q4:} Does Frame-wise Segment Refinement (FWSR) enhance real-time responsiveness without degrading motion quality?
\end{itemize}

\subsection{Experimental Setup}

\textbf{Task and Datasets.} We evaluate ReMoGen in a real-time interaction-to-reaction setting, where at each timestep, the model observes past context and autoregressively generates the entire future motion sequence. The universal single-person prior is pretrained on HumanML3D~\cite{Guo2022HumanML3D}. The Meta-Interaction Modules are trained independently: the HHI module on Inter-X~\cite{Xu2024InterX}, and the HSI module on LINGO~\cite{Jiang2024Autonomous}. Mixed-modality generalization and fine-tuning efficiency are tested on EgoBody~\cite{zhang2022egobody}, which features mixed human-human-scene interactions. All sequences are downsampled to 10 FPS in dataset preprocessing.

\textbf{Metrics.} We report standard metrics for realism and semantic alignment (FID, R-Precision, MM-Dist.), Diversity, smoothness (Peak Jerk), Collision(\%)
  and per-frame inference latency under online autoregressive rollout. We provide more detailed contact-based metrics in the supplementary material.

\textbf{Implementation Details.} The prior uses a latent autoregressive diffusion backbone, trained on HumanML3D with a history length \(H=2\), future length \(F=8\), and 10 diffusion steps. The HHI and HSI modules are trained for 65k iterations each on Inter-X and LINGO, respectively, while the prior remains frozen. The Frame-wise Segment Refinement (FWSR) module is trained for an additional 65k steps, with all other components frozen. We use AdamW (\(1 \times 10^{-4}\) learning rate), a batch size of 1024, gradient clipping (1.0), and EMA (0.999). All experiments are conducted on a single NVIDIA RTX 3090 GPU.

\subsection{Comparison with State-of-the-Art Methods}

\paragraph{Human-Human Interaction (HHI).}
On the Inter-X dataset, ReMoGen demonstrates superior performance compared to specialized HHI methods, as shown in Tab.~\ref{tab:hhi}. We compare against \textbf{ReGenNet}~\cite{Xu2024ReGenNet} (action-to-reaction), \textbf{SymBridge}~\cite{Chen2024SymBridge} (real-time Human-Robot Interaction), and \textbf{FreeMotion}~\cite{Fan2024FreeMotion} (diffusion-based). For a fair comparison, FreeMotion is trained under the same configuration as our model. ReMoGen outperforms all baselines across nearly all metrics, including FID, R-Precision, and MM-Dist, indicating higher realism and semantic accuracy. Crucially, it achieves this while maintaining a low per-frame latency (0.042s), successfully avoiding the fidelity-latency trade-off common in other real-time systems. This result affirms that our prior-guided modular design enables high-quality, real-time reaction generation.

\begin{table*}[h]
\centering
\begin{tabularx}{\textwidth}{XCCCCCCCC}
\toprule
Method            & FID $\downarrow$           & \makecell{R-Precision\\ \small{(Top-3)} $\uparrow$ }    & MM Dist.  $\downarrow$      & Diversity$\rightarrow$      & Peak Jerk  $\downarrow$ & Collision(\%)$\downarrow$ & \makecell{Latency \\ \small{(s/frame)}$\downarrow$} \\ \midrule
GT      & 0.000 & 0.609          & 1.954          & 7.678     &  0.161     & --            & -- \\\midrule
TRUMANS~\cite{Jiang2024TRUMANS} &   4.731    &      0.178          &    10.822           &      6.543            &     0.279     &   10.754    &  \textcolor{mygreen}{0.074}        \\
LINGO~\cite{Jiang2024Autonomous}   &   3.633    &         0.218       &       9.597         &      5.503    &   0.252         &   10.610  &   \textcolor{myred}{0.189}        \\\midrule
Ours    &  \textbf{1.201} & \textbf{0.530} & \textbf{3.408} &  \textbf{7.421}     &  \textbf{0.169}     &  \textbf{7.822} & \textcolor{mygreen}{0.042}    \\ \bottomrule       
\end{tabularx}
\caption{Human-Scene Interaction results on LINGO. 
\textbf{Bold} indicates best performance. \textcolor{mygreen}{Green} latency values meet the 0.1\,s/frame real-time threshold. Ours outperforms both baselines across all evaluation metrics.}
    \label{tab:hsi}
\end{table*}
 
\begin{table}[h]
\resizebox{\linewidth}{!}{
\begin{tabular}{ccccc}
\toprule
Type & Step & FID $\downarrow$           & MM Dist. $\downarrow$       & Div. $\rightarrow$  \\ \midrule
\multicolumn{2}{c}{\color[HTML]{656565} GT} & \color[HTML]{656565} 0.000          & \color[HTML]{656565} 2.894 & \color[HTML]{656565} 1.399        \\ \midrule

HHI & Zeroshot  & 18.377          & 8.665        & \textbf{3.140}       \\
HSI & Zeroshot  & 15.087          & 8.669        & 4.360           \\
Compose & Zeroshot  & \textbf{14.740}          & \textbf{8.333}        & 4.557       \\
\midrule
From Scratch & 500k &  2.341         & 4.474         & 2.718  \\
\midrule
\multirow{4}{*}{Use Prior} & 2k    & 2.757          & 4.479        &   2.908       \\
 & 5k   & 0.569          & 3.536       & 2.038        \\
 & 10k  & 0.418         & 3.413        & 1.867         \\
 & 65k    &  \textbf{0.292}         & \textbf{3.040}          & \textbf{1.429}            \\

\bottomrule
\end{tabular}
}
\caption{Mixed-modality results on EgoBody. Zero-shot composition provides complementary priors but reflects clear domain mismatch. Prior-initialized finetuning rapidly adapts to the target domain, outperforming scratch training within a few thousand steps.}
    \label{tab:egobody}
\end{table} 

\paragraph{Human-Scene Interaction (HSI).}
In the HSI task on the LINGO dataset, ReMoGen again surpasses strong, scene-aware baselines. As detailed in Tab.~\ref{tab:hsi}, we outperform both \textbf{TRUMANS}~\cite{Jiang2024TRUMANS} and \textbf{LINGO}~\cite{Jiang2024Autonomous}, which use the same segment-based rollout protocol and voxelized scene representation. ReMoGen not only achieves the \textbf{best performance across all metrics} but also records the \textbf{lowest latency} (0.042s), comfortably meeting the 0.1 s real-time threshold that one baseline fails to meet. These results validate that our Meta-Interaction Modules effectively condition generation on interaction cues (\textbf{Q2}) without compromising real-time efficiency.

\subsection{Generalization to Unseen Mixed-Modality Interaction Settings}

On EgoBody (Tab.~\ref{tab:egobody}), which contains mixed HHI and HSI scenarios unseen during training, using either the HHI or HSI module alone in a zero-shot manner yields poor performance due to the domain mismatch. However, a simple symmetric composition (\(\alpha_{\text{HHI}} = \alpha_{\text{HSI}} = 0.5\)) improves FID and MM-Dist. Initializing from our universal prior and fine-tuning on EgoBody accelerates adaptation. With 2k-10k steps, we already surpass a from-scratch model trained for 500k steps. By 65k steps, its performance approaches ground-truth statistics.

\begin{table}[]
\resizebox{\linewidth}{!}{
\begin{tabular}{cccccc}
\toprule
Method & FID $\downarrow$           & R-Prec. $\uparrow$   & MM Dist. $\downarrow$       & Div. $\rightarrow$     \\ \midrule
\color[HTML]{656565} GT    & \color[HTML]{656565} 0.000          & \color[HTML]{656565} 0.472         & \color[HTML]{656565} 4.051          & \color[HTML]{656565} 4.084                  \\ \midrule

Prior Only     & 3.735          & 0.231        & 5.736        & 3.235         \\
Scratch (No Prior)     & 0.270         & 0.412         & 4.385         & 3.907         \\
Joint-Finetune  &  0.298         & 0.439          & 4.188        & 3.862        \\
Ours  & \textbf{0.181}          & \textbf{0.464}         & \textbf{4.076}          & \textbf{3.911}         \\
\bottomrule
\end{tabular}}
\caption{Ablation on different ways of using the universal prior.}
    \label{tab:abla_2stage}
\end{table} 

\subsection{Ablation Studies}
We conduct targeted ablations to validate the main components of our framework.

\paragraph{Impact of the Universal Prior.}
To answer \textbf{Q1}, we analyze the contribution of our pretrained universal prior in Tab.~\ref{tab:abla_2stage}. A model using the \textbf{Prior Only} struggles to generate coordinated interactive motion, confirming a distribution mismatch. Training a model \textbf{from Scratch} on Inter-X alleviates this but suffers from unstable kinematics due to the limited dataset size. \textbf{Joint-Finetuning} of the entire model improves semantic alignment but degrades motion quality by eroding the rich kinematic knowledge from the large-scale pretraining.

In contrast, our approach of keeping the universal prior \textbf{frozen} and adapting via Meta-Interaction Modules strikes the optimal balance. It preserves the strong, generalizable motion structures from the prior while enabling targeted adaptation to interaction cues. This strategy is key to achieving superior realism and semantic consistency.

\paragraph{Efficacy of Frame-wise Segment Refinement (FWSR).}
To address \textbf{Q4}, we evaluate the impact of FWSR on responsiveness in Tab.~\ref{tab:abla_fwr}. The standard \textbf{segment-based rollout} is real-time but has limited reactivity, as updates occur only once per segment. Conversely, a naive \textbf{frame-wise diffusion} approach (Slide) reacts instantly but incurs prohibitive latency and introduces temporal artifacts.

FWSR resolves this trade-off. By applying a lightweight, per-frame refinement adapter on top of a stable segment-level prediction, it injects fine-grained adjustments based on the latest interaction cues. This design significantly improves responsiveness to dynamic events while keeping runtime nearly constant, thereby enhancing real-time interaction without sacrificing the temporal stability provided by the powerful backbone generator.

\definecolor{lightgrey}{HTML}{EFEFEF}
\begin{table}[]
\resizebox{\linewidth}{!}{
\begin{tabular}{cccccccc}
\toprule
Train & Infer & FID $\downarrow$           & R-Prec. $\uparrow$   & MM Dist. $\downarrow$       & Div. $\rightarrow$     & Latency $\downarrow$     \\ \midrule
\multicolumn{2}{c}{\textcolor{mygrey}{GT}}    & \textcolor{mygrey}{0.000}          & \textcolor{mygrey}{0.472}         & \textcolor{mygrey}{4.051}          & \textcolor{mygrey}{4.084}         & \textcolor{mygrey}{--}          \\ \midrule
\rowcolor[HTML]{EFEFEF} 
\textit{Seg.} & \textit{Seg.}     & \textit{0.181}          & \textit{0.464}        & \textit{4.076}        & \textit{3.911}         &      \textit{\textcolor{mygreen}{0.042}}   \\
\textit{Slide} & \textit{Slide}     & 4.136         & 0.289         & 5.519         & 3.700         &   \textcolor{myred}{0.305} \\
\textit{Seg.} & \textit{Slide}     &  1.785         & 0.380          & 4.789        & 3.832          &   \textcolor{myred}{0.305} \\
\rowcolor[HTML]{FFFFC7} 
\textit{Seg.} & \textit{FWSR}     & \textbf{0.166}          & \textbf{0.462}         & \textbf{4.076}          & \textbf{4.003}          &   \textcolor{mygreen}{0.047} \\
\bottomrule
\end{tabular}}
\caption{Ablation on Frame-wise Segment Refinement (FWSR). \textit{Seg.} denotes standard segment-based autoregressive rollout, which updates once per segment.
\textit{Slide} regenerates a full segment at every frame and uses only the first predicted frame.
The gray row shows the baseline segment rollout, and the highlighted row shows our FWSR, which performs per-frame refinement while keeping segment-level generation unchanged.}
    \label{tab:abla_fwr}
\end{table}
\section{Conclusions}
We introduced ReMoGen, a modular framework for real-time interaction-to-reaction motion generation that combines a universal single-person motion prior, Meta-Interaction Modules, and an efficient Frame-wise Segment Refinement mechanism. By decoupling universal motion dynamics from domain-specific interaction cues, ReMoGen leverages modular learning to adapt data-efficiently across diverse HHI and HSI datasets, generalizes to unseen mixed-interaction scenarios, and rapidly specializes to new environments. The frame-wise refinement mechanism further improves responsiveness under dynamic conditions while preserving the temporal coherence of segment-based generation. Extensive experiments show that ReMoGen produces high-quality, semantically aligned, and real-time reactive motion across diverse interaction scenarios, providing a scalable foundation for downstream applications.
\section{Acknowledgment}
This work was supported by MoE Key Laboratory of Intelligent Perception and Human-Machine Collaboration (KLIP-HuMaCo), Shanghai Frontiers Science Center of Human-centered Artificial Intelligence (ShangHAI).
{
    \small
    \bibliographystyle{ieeenat_fullname}
    \bibliography{main}
}
\clearpage
\setcounter{page}{1}
\setcounter{section}{0}
\setcounter{figure}{0}
\maketitlesupplementary
\thispagestyle{empty}
\appendix

\setcounter{figure}{0}
\setcounter{table}{0}
\renewcommand{\thefigure}{\Alph{figure}}
\renewcommand{\thetable}{\Alph{table}}


\section{Overview}
\label{sec:overview}

This supplementary material provides a detailed description of the
\emph{ReMoGen} framework for real-time human interaction-to-reaction
generation. Due to space constraints, the main paper only briefly
summarizes a number of design choices and implementation details. Here
we expand all core components so that the framework is fully
specified and reproducible. 

\section{Representations}
\label{sec:notation}

\subsection{Motion Representation}

We adopt a SMPL-X-based motion representation aligned with our backbone design~\cite{Zhao2024DART}. 
For each frame $t$, we construct a compact feature vector $m_t$ using physically meaningful quantities derived from the root motion and the first 22 SMPL-X body joints. 
Specifically, we extract:
\begin{itemize}
    \item Root translation $\mathbf{t}_t$ in the canonical frame.
    \item Joint rotations $\mathbf{R}_t$ in continuous 6D form for the root and body joints.
    \item Root motion deltas, including translation changes $\Delta \mathbf{t}_t$ and orientation changes $\Delta \mathbf{R}_t$ between consecutive frames.
    \item Joint positions $\mathbf{J}_t$ obtained from the SMPL-X forward pass.
    \item Joint velocities $\Delta \mathbf{J}_t$ computed as per-frame differences.
\end{itemize}

The final per-frame representation $m_t$ is formed by concatenating all components above, and is used throughout the training and inference stages of our autoregressive latent-diffusion backbone.

\paragraph{Canonicalization and Normalization.}

Before computing features, all SMPL-X sequences 
$\{ \mathbf{t}_t, \mathbf{R}_t, \mathbf{J}_t \}_{t=1}^{T}$
are transformed into a consistent \emph{ego-centric canonical frame}.

The pelvis joint of the first frame defines the canonical origin $\mathbf{p}_0$.

The body facing direction is estimated from the left and right hip joints (e.g., joints $j_1$ and $j_2$), which determines the canonical horizontal axes.

A rigid transform $(\mathbf{R}_{\text{ego}}, \mathbf{t}_{\text{ego}})$ is then constructed from this reference frame and applied to the root translation $\mathbf{t}_t$, root orientation $\mathbf{R}_t$, and joint positions $\mathbf{J}_t$ of the entire sequence, producing the canonicalized quantities 
$\tilde{\mathbf{t}}_t$, 
$\tilde{\mathbf{R}}_t$, 
and 
$\tilde{\mathbf{J}}_t$.

For human-human interaction scenarios, both the ego and the partner sequences are transformed using the same ego-defined frame, ensuring that all agents' trajectories are expressed consistently relative to the canonicalized ego.

After canonicalization, we compute the global mean and standard deviation of all feature channels using only the training split and subsequently normalize every motion tensor.

\paragraph{History and Future Segmentation.}

We operate in a segment-based autoregressive regime. Each internal
step $i$ of the generator operates on:
\begin{itemize}[leftmargin=*]
  \item a history window
  $M_{h}^{i} \in \mathbb{R}^{H \times D}$ of length $H$, and
  \item a future segment
  $\hat{M}_{f}^{i} \in \mathbb{R}^{F \times D}$ of length $F$.
\end{itemize}

After generating the $i$-th future segment, we update the next history
by sliding a window of size $H$:
\begin{equation}
  M_{h}^{i+1}
  =
  \big[\text{concat}(
    M_{h}^{i}, \hat{M}_{f}^{i})
  \big]_{-H:}.
  \label{eq:history_update}
\end{equation}
This corresponds to the update rule described in the main paper.
\subsection{Scene Representation}
Our framework supports human-scene and human-human-scene interactions by encoding the environment as a volumetric voxel occupancy grid~\cite{Jiang2024Autonomous,Jiang2024TRUMANS}. 
We adopt a unified 3D representation across all scene-aware datasets.

\paragraph{Voxel Occupancy.}
For datasets that natively provide voxelized scenes (e.g., LINGO), we directly use their occupancy volumes.
Each scene is represented as a dense grid of size \textbf{$x \times y \times z$} with a spatial voxel resolution $v_{size}$. 
The voxel grids are loaded and cached in the dataset loader, and queried during training using point-based lookup.

\paragraph{Ego Voxel Representation.}
The Scene Encoder receives an \emph{ego-centric voxel occupancy} volume that describes the local spatial layout around the moving subject. 
The ego voxel grid is defined over a fixed bounding box:
\[
x \in [-0.6,\, 0.6], \quad 
y \in [-0.6,\, 0.6], \quad
z \in [0.1,\, 1.2],
\]
resulting in a compact \textbf{$32 \times 32 \times 32$} occupancy volume. 

This bounding box corresponds to a localized workspace around the body. To construct ego voxels, we query the underlying scene occupancy grid at the projected 3D locations of the root joint (or goal positions). 
\subsection{Ego Alignment.}
All motion features are represented in our ego-centric canonical coordinate frame for consistent modeling across datasets. 
Before being fed into the network, any other motions or goal positions associated with the ego are also transformed into this canonical frame, ensuring that the model interprets all positional objectives relative to the ego-centered coordinate system.

However, scene occupancy queries must be performed in the global scene coordinate system. 
Thus, although the model operates in the canonical space, we temporarily convert the canonicalized ego motion back to world coordinates using the stored inverse canonical transform whenever querying the ego voxel grid.

The reconstructed global root positions are then used to sample occupancy values from the corresponding voxel volumes. 
This two-space design maintains consistent interaction between ego motion, partner motion, and scene geometry while enabling stable canonical-frame learning inside the model.

\section{Supplement Method Details}
\label{sec:method_details}

\subsection{Meta-Interaction Modules}
\label{subsec:meta_interaction}
\begin{figure}[h]
  \centering
  \includegraphics[width=\linewidth]{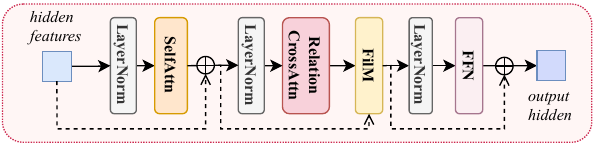}
   \caption{Architecture of Meta-Interaction Block.}
   \label{fig:inter_module}
\end{figure}

The universal prior does not model interactions by construction. To
inject dynamic context from other agents and the scene while keeping
the prior frozen, we introduce \emph{Meta-Interaction Modules} (MIMs).
Each MIM is an adapter that modulates intermediate activations in the
latent denoiser via FiLM-style conditioning on external context.
\paragraph{Meta-Interaction Block.}
A Meta-Interaction Block (Fig.~\ref{fig:inter_module}) receives ego features 
$h \in \mathbb{R}^{B \times T \times d}$ 
and context tokens 
$c_{\text{context}} \in \mathbb{R}^{B \times T_c \times D}$ 
(from other agents or from the scene encoder).
The block outputs a context-induced residual that is added to the ego features.

\paragraph{Self-Attention.}
We first apply standard self-attention to capture intra-sequence dependencies:
\begin{equation}
    h' = h + \operatorname{SelfAttn}(\operatorname{LayerNorm}(h)).
\end{equation}

\paragraph{Relation-aware Cross-attention.}
Given contextual embeddings $c_{\text{context}}$, we compute cross-attention using $ Q_h = \operatorname{LayerNorm}(h') W_Q, K_c = c_{\text{context}} W_K, V_c = c_{\text{context}} W_V $
where $W_Q, W_K, W_V$ are learned projections.

The relation-aware cross-attention output is
\begin{equation}
    r = \mathrm{Softmax}\!\left(
        \frac{Q_h K_c^\top + B_{\text{rel}}}{\sqrt{d}}
    \right) V_c ,
\end{equation}
where 
$B_{\text{rel}} \in \mathbb{R}^{T \times T_c}$ 
is a learnable relative positional bias encoding temporal or spatial offsets.

Each bias element is computed using a sinusoidal embedding of the relative index $\Delta t = i - j$:
\begin{equation}
    b_{ij} = 
    W_b\!\left[
        \sin(\omega \Delta t),\;
        \cos(\omega \Delta t)
    \right],
\end{equation}
with a fixed frequency $\omega = 0.25$ and a learned projection $W_b$.

\paragraph{FiLM-style Modulation.}
The cross-attention output $r$ is mapped to FiLM parameters \((\gamma, \beta)\)via a linear layer.
We modulate $h'$ using
\begin{equation}
    h_{\text{mod}}
    =
    (1 + \tanh \gamma) \odot h'
    +
    \tanh \beta.
\end{equation}

\paragraph{Feed-Forward Network (FFN).}
A transformer-style FFN refines the modulated features:
\begin{align}
    h_{\text{ffn}}
    &= h_{\text{mod}}
    + \text{FFN}(\operatorname{LayerNorm}(h_{\text{mod}})),
\end{align}

The final context-induced residual is
\begin{equation}
    \Delta_{\text{context}} = h_{\text{ffn}} - h.
\end{equation}
This residual is injected into the denoiser at predefined transformer layers, enabling interaction-aware modulation.

\subsection{Compose MIMs During Inference}
\paragraph{Composable Inference.}
Multiple Meta-Interaction Modules (e.g., HHI and HSI branches) can be activated simultaneously. 
At each injection layer, the system first produces per-branch residuals 
$\Delta_i \in \mathbb{R}^{B \times T \times d}$, 
such as $\Delta_{\text{others}}$ and $\Delta_{\text{scene}}$.
Their contributions are combined through a weighted sum:
\begin{equation}
\Delta_{\text{total}} = \sum_i \alpha_i\, \Delta_i ,
\label{eq:compose_sum}
\end{equation}
where each coefficient $\alpha_i$ controls the strength of a module.  
These weights may be \textbf{set manually by the user} or \textbf{predicted automatically by an external model}, 
such as a task-specific rule system or a large language model that interprets textual instructions.

\paragraph{L2-Norm Clamping for Stability.}
A naïve weighted sum can produce overly large residuals, especially when multiple modules reinforce each other. 
To ensure stable interaction conditioning while preserving the frozen prior's motion manifold,
we normalize $\Delta_{\text{total}}$ using a \emph{per-sample L2 norm clamp}.

Let 
\[
\|\Delta_{\text{total}}\|_2 = 
\left\|\Delta_{\text{total}}\right\|_{2,\text{flatten}},
\]
where the norm is computed over the $(T \times d)$ dimensions for each sample.  
Similarly, let $m$ be the largest norm among the individual modules:
\[
m = \max_i \left\|\Delta_i\right\|_{2,\text{flatten}} .
\]

We compute a scale factor
\begin{equation}
s = \min\!\left(1,\;
\frac{m}{\|\Delta_{\text{total}}\|_2 + \varepsilon}
\right),
\end{equation}
with $\varepsilon$ a small constant for numerical stability.  
The fused modulation is then:
\begin{equation}
\Delta_{\text{final}} = s \cdot \Delta_{\text{total}} .
\label{eq:l2_clamp}
\end{equation}

\paragraph{Effect of the Clamp.}
This operation guarantees that the fused residual never exceeds the magnitude of the strongest individual branch, preventing runaway updates and ensuring that the decoded motion remains on the prior's learned manifold.  
The model therefore supports \textbf{arbitrary combinations} of HHI and HSI cues—including user-driven, programmatic, or LLM-derived scaling—while maintaining temporal stability and preserving the geometry of the pretrained motion prior.

\subsection{Frame-wise Segment Refinement (FWSR)}
\label{subsec:fwsr}

Segment-based autoregressive rollout provides strong long-horizon stability, but the model updates its reaction only once per segment, introducing latency.  
To improve responsiveness under real-time constraints, we adopt a lightweight \emph{Frame-wise Segment Refinement (FWSR)} module that applies per-frame latent corrections without re-running the full diffusion process.  
FWSR operates directly on the clean VAE latent $z_0 \in \mathbb{R}^{d_z}$ predicted for the segment by the universal prior.

\paragraph{FWSR Module Overview.}
The FWSR module receives the clean latent $z_0$, the ego-dynamic feature sequence $M_h$, and dynamic context $X_{dyn}$ such as other-agent motion. 
Dynamic context tokens are first processed by a lightweight refinement block consisting of a projection layer and a self-attention encoder to obtain $c_{dyn}$.
The processed dynamic tokens, together with $M_h$, are then fed into a relation-aware cross-attention layer followed by a FiLM head. 
The architectural components mirror the Meta-Interaction Modules (MIM), but operate directly in the decoder latent space.

\paragraph{Relation-Aware Attention.}
At each refinement step, the module constructs relation features jointly from the ego-dynamic features $M_h$ and each processed dynamic context token $c_{dyn}$.  
Let
\[
Q = z_0 W_Q,\qquad  
K = c_{dyn} W_K,\qquad 
V = c_{dyn} W_V,
\]
where $W_Q, W_K, W_V \in \mathbb{R}^{d_z \times d_z}$ are learned projections.  
Relation features $(M_h, X_{dyn})$ are used to compute relative positional bias $
B_{\text{rel}}$.  
Cross-attention is then computed as:
\[
\Delta_{\text{attn}}
=
\operatorname{RelAttn}\!\left(
    Q,\;
    K,\;
    V,\;
    \mathrm{RelBias}(M_h, X_{dyn})
\right),
\]
allowing the ego latent to incorporate fine-grained, interaction-dependent refinements.

\paragraph{FiLM Latent Modulation.}
The attention output is mapped to FiLM parameters
$(\gamma, \beta)$, 
and applied to refine the latent:
\[
\Delta_{\text{FWSR}}
=
(1+\tanh \gamma)\odot z_0 + \tanh \beta - z_0.
\]

\paragraph{Safe Latent Scaling via Decoder Sensitivity.}
Directly adding $\Delta_{\text{FWSR}}$ may push the latent into directions where the decoder is overly sensitive.  
To stabilize the refinement, we use a \emph{decoder sensitivity vector} 
$s \in \mathbb{R}^{d_z}$, 
estimated using finite-difference gradients of the decoder with respect to each latent dimension.

We apply per-dimension scaling:
\[
\Delta_{\text{safe},d}
=
\frac{\Delta_{\text{FWSR},d}}{1 + \beta s_d},
\]
where $\beta$ is a hyper-parameter controlling suppression strength. We use $\beta=1.0$.  
This protects highly sensitive latent dimensions, suppressing artifacts while leaving robust directions expressive.

\paragraph{Final Latent Refinement.}
The refined latent is obtained as:
\[
\tilde{z} = z_0 + \Delta_{\text{safe}}.
\]
FWSR performs this refinement once per frame at negligible cost, enabling low-latency adjustments and improved responsiveness while keeping all backbone and interaction modules frozen.

\section{Experimental Setup Details}
\label{sec:datasets}
\subsection{Datasets and Pre-processing}
We summarize datasets used for different stages and the pre-processing
steps that standardize them into a unified representation.
\paragraph{HumanML3D.}
The \textbf{Universal Single-Person Prior} is trained on HumanML3D~\cite{Guo2022HumanML3D}. We use the official train/val/test splits and its original text annotations.

\paragraph{Inter-X.}
The Meta-Interaction Module for Human-Human Interaction is trained on Inter-X~\cite{Xu2024InterX}, which contains paired SMPL-X sequences of two interacting subjects. We use its official train/val/test splits.

To increase coverage and balance role diversity, we perform data augmentation by treating \textbf{both \texttt{p1} and \texttt{p2} as the ego} in separate passes. 
For each choice of ego, the remaining person is treated as the conditioning partner, and we construct history-future motion segments for the ego while using the partner's motion as dynamic context.

Inter-X provides only \emph{global} textual descriptions for each two-person interaction, without distinguishing the roles of each person. 
To obtain ego-specific conditioning texts, we use the DeepSeek V3 large language model to automatically decompose each original dual-person description into two separate role-specific descriptions. 

We supply the LLM with the raw Inter-X action label and multiple dataset-provided textual descriptions, and instruct the model to identify the actor and reactor and rewrite the interaction into two independent third-person declarative sentences follows the style of HumanML3D. 
The final assignment of actor/reactor descriptions to \texttt{p1} or \texttt{p2} follows the dataset's official \texttt{interaction\_order} annotation.
This process allows Inter-X's coarse two-person textual descriptions to be converted into well-aligned, role-specific conditioning signals for training our HHI Meta-Interaction Module.

\paragraph{LINGO.}
The Meta-Interaction Module for Human-Scene Interaction is trained on LINGO~\cite{Jiang2024Autonomous}, which provides SMPL-X motions paired with detailed 3D indoor scenes. 
We use the dataset's original text annotations as the semantic descriptions for each motion sequence without additional rewriting.

For scene representation, we follow the Z-up-aligned official LINGO protocol and use the provided \emph{voxelized occupancy grids}. 
Each indoor environment is represented as a dense occupancy volume defined over the following axis-aligned bounds:
\[
x \in [-3.0,\, 3.0], \quad
y \in [-4.0,\, 4.0], \quad
z \in [0.0,\, 2.0],
\]
which is discretized into a grid of size $\textbf{300} \times \textbf{400} \times \textbf{100}$ with a spatial resolution of \textbf{0.02\,m}. 
This large scene-centric voxel structure captures room-scale spatial layouts, obstacles, and affordances and is used as the conditioning input for our scene encoder.

Since LINGO does not release ground-truth motions for a held-out test set, we construct our own splits by randomly partitioning the available training data into \textbf{70\% training}, \textbf{10\% validation}, and \textbf{20\% testing}. 
All reported HSI results in our paper are based on this split.

\paragraph{EgoBody.}
We use EgoBody~\cite{zhang2022egobody} to evaluate Human-Human-Scene generalization and compositionality. 
For textual conditioning, we adopt the \textbf{sequence-level motion descriptions} provided by Motion-X, which offers annotations aligned with each full interaction sequence. 

To maintain consistency with our HHI preprocessing, we apply the same data augmentation strategy used in Inter-X by \textbf{swapping the ego and the other agent}. 
In each augmented version, the chosen ego is canonicalized and the partner is treated as the conditioning sequence.

EgoBody only provides scene geometry in \textbf{mesh format}. 
To convert the scenes into a unified voxel representation, we first normalize each scene by translating the mesh such that its \textbf{XY center lies at the $(0,0)$} and its \textbf{lowest Z coordinate aligns with the ground plane} $(z=0)$. 
We then voxelize the normalized mesh using the \textbf{same occupancy configuration as LINGO}.

\paragraph{General Pre-processing.}

For all datasets, we directly use the provided SMPL-X parameters and convert motions into a consistent \textbf{Z-up} coordinate convention. All SMPL-X models are loaded with the \texttt{neutral} gender configuration for consistency across datasets. We then obtain body joints by forwarding the SMPL-X parameters through the neutral-gender model.

For all ego sequences, we discard the dataset-provided body shape and set $\boldsymbol{\beta} = \mathbf{0}$ to eliminate cross-subject shape variation.  
For interaction scenarios involving other agents, we retain the original SMPL-X \(\boldsymbol{\beta}\) parameters for the partner(s) to preserve their identity and body proportions as provided in each dataset.

All datasets are uniformly downsampled to \textbf{10 FPS} using equal-interval sampling to match the real-time generation setting of our framework.

For single-person datasets (HumanML3D) and human-human interaction datasets (Inter-X), we shift the mesh vertically such that the lowest vertex of the first-frame mesh lies exactly on the ground plane.  
For interaction datasets that include scenes (LINGO, EgoBody), we \emph{do not modify} the absolute spatial placement; the original world-coordinate positions and contact relationships with the environment are strictly preserved.

\subsection{Training Details}
\label{sec:training_details}

Our model is trained in a progressive manner that first learns a universal motion prior, then incorporates interaction-specific conditioning through Meta-Interaction Modules, and optionally adds a Frame-wise Segment Refinement module for enhancing responsiveness to dynamic contexts. 

\paragraph{Universal Single-Person Motion Prior.}
We begin by training a universal motion prior on HumanML3D under a purely single-person setting. 
The model predicts future motion segments using an autoregressive latent diffusion process with a history window of $H = 2$ frames and a prediction length of $F = 8$ frames. 
We adopt a DDPM-based denoising formulation with \textbf{10 diffusion steps}.
All feature \textbf{mean and standard deviation statistics} used for normalization are computed exclusively during this stage and \textbf{reused unchanged} across all subsequent interaction datasets (Inter-X, LINGO, EgoBody), ensuring consistent feature scaling throughout training. 
This phase consists of \textbf{250k} steps for VAE training and \textbf{400k} steps for training the diffusion denoiser.

\paragraph{Meta-Interaction Learning.}
Once the universal prior is learned, we introduce Meta-Interaction Modules that incorporate external interaction cues such as other agent's motion (Inter-X) or scene occupancy (LINGO). 
During this process, the universal backbone remains \emph{frozen}. 
Each Meta-Interaction module is trained for \textbf{65k} steps using its respective dataset.

\paragraph{Frame-wise Segment Refinement (FWSR).}
This module refines autoregressive predictions through per-frame residual updates while keeping both the universal prior and Meta-Interaction parameters fixed. 
As FWSR is not required for training the core interaction model, it is applied only when there are dynamic contexts like others motion. 
The refinement module is trained for \textbf{65k} steps.

\subsection{Evaluation Protocol and Metrics}
\label{subsec:eval}

\paragraph{Evaluation Protocol.}
We evaluate all models in an \emph{Online Interaction-to-Reaction} setting, where the ego agent continuously observes interaction cues (other-agent motion or scene occupancy) and produces an immediate reaction. 
To ensure fair comparison, we strictly follow each baseline's original inference procedure:
\begin{itemize}[leftmargin=*]
    \item \textbf{Baselines with autoregressive rollout.}  
    If a baseline provides an online or autoregressive generation mechanism, we use it unchanged. 
    The model receives the same history window as specified by its official implementation, without adding any auxiliary history embeddings.
    \item \textbf{Baselines without autoregressive rollout.}  
    If a baseline does not support online generation, we follow the “segment prediction and stitching” protocol: the model predicts motion in fixed-length segments, and consecutive segments are stitched together by recursively feeding the end of the previous segment into the next prediction window. 
    No modifications or additional temporal encodings are introduced beyond what the baseline originally supports.
    \item \textbf{Models without history conditioning.}  
    For baselines that do not learn history embeddings or do not accept temporal context explicitly, we do \emph{not} introduce any additional history mechanism.  
    They receive only the inputs allowed by their native design.
\end{itemize}

All evaluations are conducted at 10\,FPS after canonicalization and standardization, consistent with our training setup.

\paragraph{Metrics.}
We adopt a comprehensive set of quantitative metrics widely used in human motion synthesis and interaction modeling. 
These metrics cover semantic alignment, distributional realism, motion diversity, physical smoothness, and real-time system performance.

\textbf{Embedding-based Metrics.}
Following standard practice in HumanML3D, we extract normalized motion and text features using the pretrained t2m motion and language encoders~\cite{Guo2022HumanML3D}.  
The following metrics are computed in this embedding space:

\begin{itemize}[leftmargin=*]
    \item \textbf{FID.} compares the embedded feature distributions of generated and ground-truth motions, measuring distributional realism. Lower FID indicates that the generated motion better matches the statistics of natural human movement.
    
    \item \textbf{R-Precision.} evaluates semantic alignment by retrieving the correct textual description of a motion among a batch of 64 candidates. We report top-3 accuracy, reflecting how often the ground-truth description ranks among the three most similar texts based on cosine similarity.
    
    \item \textbf{Diversity.} quantifies variation among generated motions by computing the average embedding-space distance between multiple samples, indicating the model's ability to avoid mode collapse.
    
    \item \textbf{MM-Dist.} measures motion-text semantic consistency by computing the embedding-space distance between a generated motion and its paired textual description. Lower MM-Dist. corresponds to stronger alignment between generated motion and intended semantics.
\end{itemize}

\textbf{Peak Jerk.}
This metric captures high-frequency artifacts by measuring the maximum time derivative of joint acceleration. 
Lower jerk values correspond to smoother, more physically plausible motion trajectories.

\textbf{Latency.}
We report the average per-frame computation time (in seconds) under our Online Interaction-to-Reaction protocol, measured over 1,000 generated frames.  
This reflects the model's responsiveness and suitability for real-time deployment.
\begin{table*}[t]
\centering
\begin{tabularx}{\textwidth}{XCCCC|CCC}
\toprule
Method            & FID $\downarrow$           & \makecell{R-Precision\\ \small{(Top-3)} $\uparrow$ }    & MM Dist.  $\downarrow$      & Diversity$\rightarrow$      & \makecell{Contact \\ \small{Precision} $\uparrow$ }  & \makecell{Contact \\ \small{Recall} $\uparrow$ } & Collision(\%)$\downarrow$\\ \midrule
\textcolor{mygrey}{GT}                & \textcolor{mygrey}{0.000}          & \textcolor{mygrey}{0.472}          & \textcolor{mygrey}{4.051}          & \textcolor{mygrey}{4.084}                   & \textcolor{mygrey}{--}         & \textcolor{mygrey}{--}  & \textcolor{mygrey}{--}           \\\midrule
ReGenNet~\cite{Xu2024ReGenNet}          &       11.622         &       0.269         &       6.092         &           3.452           &       0.549       & 0.505 & 0.665 \\
FreeMotion~\cite{Fan2024FreeMotion}        & 3.383          & 0.284          & 5.438          & 3.394          & 0.497          & 0.321 &  0.545          \\
\(\text{FreeMotion}^{\text{off}}\)~\cite{Fan2024FreeMotion}    & 0.492          & 0.417          & 4.330          & 3.913           &  0.496    &   0.484 &     1.220           \\
SymBridge~\cite{Chen2024SymBridge} & 2.569          & 0.355          & 4.955          & 3.598                 & 0.614  &    0.447 & 0.646      \\ \midrule
Ours              & 0.181         &0.464 & 4.076 & 3.911        & 0.527    &   0.415   &  0.989           \\
Ours+FWSR   &0.166 & 0.462    & 4.076 & 4.003 & 0.533     & 0.423& 0.961       \\ \bottomrule
\end{tabularx}

\caption{Evaluation with contact-based metrics on HHI.}
\label{tab:contact}
\end{table*}

\textbf{Contact-based Metrics.}
We report \emph{Collision Rate} and other contact-based metrics in Tab.~\ref{tab:contact}.  \emph{Collision Rate} is defined as the ratio of frames in which body joints intersect with the scene mesh or other agents' meshes. For other contact-based metrics, we follow SymBridge~\cite{Chen2024SymBridge}, which adopts contact-related metrics from the human-object interaction (HOI) setting. Specifically, we report \emph{Contact Precision} and \emph{Contact Recall}, which measure the alignment between predicted and ground-truth contact patterns.

\section{Additional Experimental Results}
\subsection{Contact-based Metrics Results}

We provide additional contact-based evaluation to complement the main results (Tab.~\ref{tab:contact}). ReMoGen achieves substantially better motion quality (FID) while maintaining competitive contact alignment compared to prior methods. 

We note that contact outcomes in interaction scenarios are inherently \emph{multi-modal}. Multiple distinct contact patterns can correspond to equally plausible interactions depending on timing, spatial configuration, and agent behavior. As a result, metrics that directly compare predicted contacts to a single ground-truth pattern may underestimate valid alternative interactions. 

\subsection{Latency Breakdown}

We provide a detailed breakdown of runtime (Tab.~\ref{tab:latency}) to complement the end-to-end latency reported in the main paper. The total latency corresponds to end-to-end inference time measured on a single GPU. Time not explicitly assigned to the listed components is attributed to data I/O and other implementation overhead.

\begin{table}[h]
\centering
\small
\begin{tabular}{lc}
\toprule
Component & Time (s) \\
\midrule
Denoising (total, 10 cond + 10 uncond) & 0.29 \\
\quad Denoiser & 0.067 \\
\quad Meta-Interaction Modules & 0.069 \\
\quad per step & 0.0136 \\
Decoding & 0.005 \\
Frame-wise Segment Refinement & 0.047 \\
\quad decoding & 0.005 \\
\quad FWSR module & 0.0017  \\
\quad per iteration & 0.0067 \\
Pre/Post-processing (canonicalization) & $\sim$0.02 \\
\midrule
Total (8 frame or less) & $\sim$0.36 \\
\quad per frame & $\sim$0.047 \\
\bottomrule
\end{tabular}
\caption{Latency breakdown of ReMoGen.}
\label{tab:latency}
\end{table}

\begin{figure*}[t]
    \centering
    \includegraphics[width=1.0\linewidth]{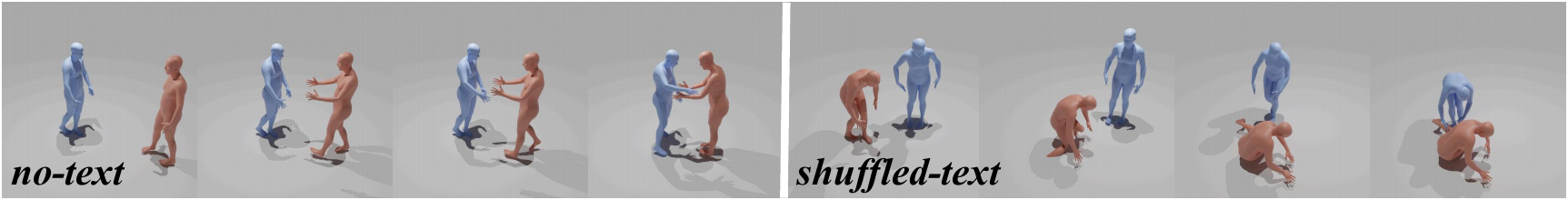}
   \captionof{figure}{Results under \emph{no-text} and \emph{shuffled-text} settings. In \emph{no-text}, the original intent is ``Shake hands''. In \emph{shuffled-text}, the original intent is ``Help others up'' while the input is ``Standing still''.}
   \label{fig:text}
\end{figure*}
\section{Robustness Analysis}
We analyze the robustness of ReMoGen along two dimensions: (i) sensitivity to semantic inputs, and (ii) sensitivity to different context encoder choices.
\label{sec:robust}
\subsection{Robustness to Semantic Inputs}

We evaluate the robustness of ReMoGen to high-level semantic inputs by testing two settings: (i) \emph{no-text}, where no semantic signal is provided, and (ii) \emph{shuffled-text}, where the input text conflicts with the physical interaction context.

As shown in Fig.~\ref{fig:text}, ReMoGen remains stable in both settings. When semantic inputs are absent or inconsistent, interaction-driven cues dominate motion generation, leading to physically plausible reactions. This behavior is consistent with our design, where text is consumed only by the universal prior, while Meta-Interaction Modules are text-agnostic and react directly to dynamic interaction cues.

\begin{table}[h]
\resizebox{\linewidth}{!}{
\begin{tabular}{ccccc}
\toprule
Encoder Type & FID $\downarrow$         & R-Precision $\uparrow$   & MM Dist. $\downarrow$  & Div. $\rightarrow$     \\ \midrule
\color[HTML]{656565} GT    & \color[HTML]{656565} 0.000          & \color[HTML]{656565} 0.472         & \color[HTML]{656565} 4.051          & \color[HTML]{656565} 4.084                  \\ \midrule
MLP   &     0.236      &    0.464   &    4.109  & 3.890   \\
Transformer   &    0.223    &      0.457    &   4.135 &   3.816 \\
TCN (Ours)  &     0.181      &     0.464     &      4.076 & 3.911   \\
\bottomrule
\end{tabular}}
\caption{Encoder sensitivity evaluation on HHI.}
    \label{tab:abla_enc}
\end{table}
\subsection{Robustness to Encoder Design}
We evaluate the sensitivity of ReMoGen to different context encoders by replacing the TCN encoder with alternative architectures, including MLP and Transformer-based encoders.

As shown in Tab.~\ref{tab:abla_enc}, almost all variants consistently outperform baselines (in Tab.~\ref{tab:hhi}), and the performance differences between encoder choices are relatively small. This indicates that the effectiveness of ReMoGen does not depend on a specific encoder design, but rather on the proposed modular framework and training strategy.

\section{Qualitative Experimental Results}
\label{sec:extra_results}
This section presents extended qualitative evaluations of ReMoGen across Human-Human, Human-Scene, and Human-Human-Scene interaction scenarios. These visualizations complement the quantitative results summarized in Tab.~\ref{tab:hhi}, Tab.~\ref{tab:hsi}, Tab.~\ref{tab:egobody}, Tab.~\ref{tab:abla_2stage}, and Tab.~\ref{tab:abla_fwr}, and further illustrate how our modular learning design achieves stable, responsive, and semantically aligned motion generation.
\subsection{Comparison Results}
\paragraph{Human-Human Interaction.}
\begin{figure*}[t]
    \centering
    \includegraphics[width=1.0\linewidth]{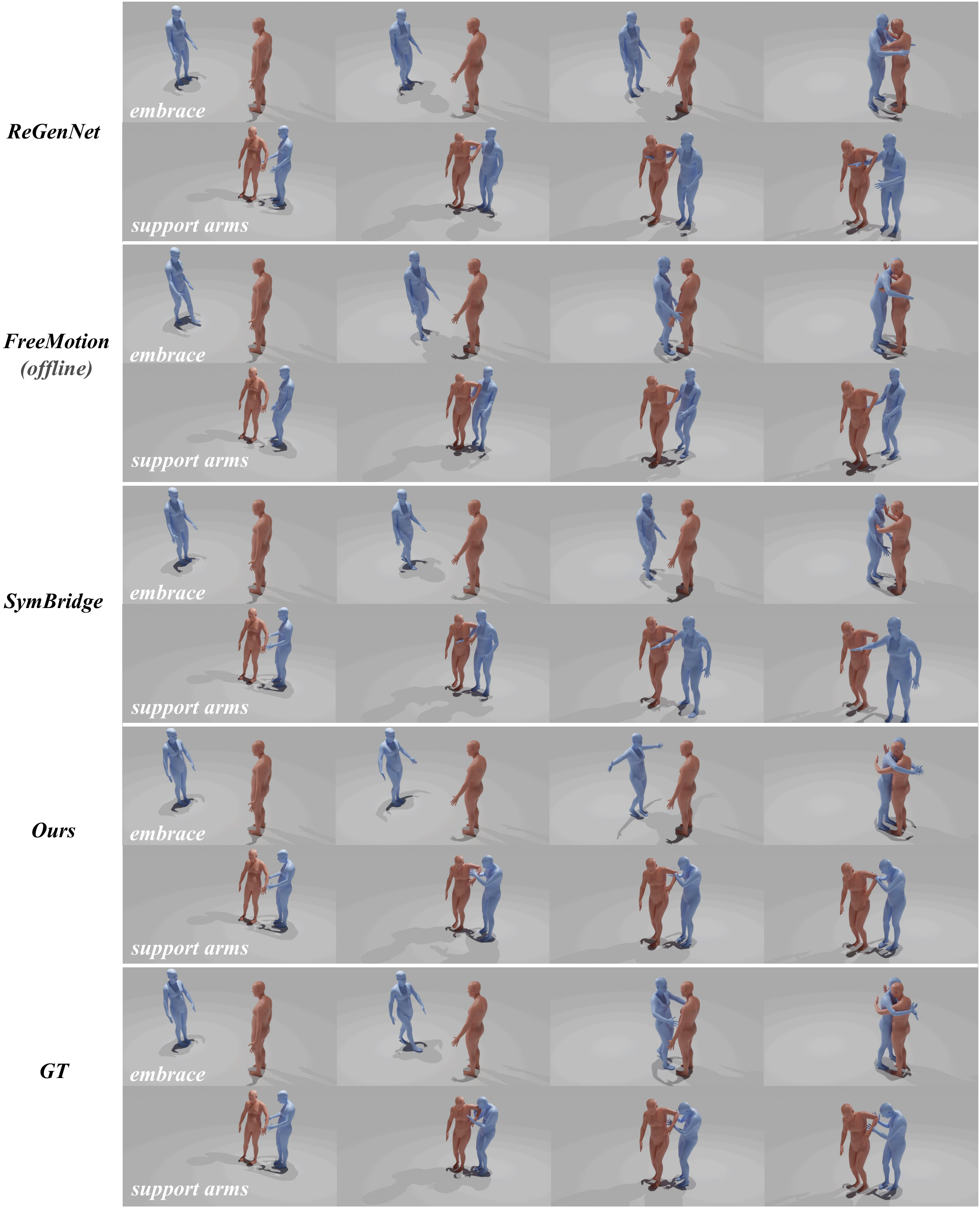}    \caption{Qualitative comparisons on Human–Human Interaction tasks. For typographical reasons, we have presented the optimal offline version of FreeMotion.
Our method produces smoother and more coordinated reactions aligned with the intent, whereas baselines exhibit unnatural timing, misaligned contact, or unstable body dynamics.}
    \label{fig:hhi}
\end{figure*}
We present two representative examples: \textit{“One person approaches and opens their arms to \textbf{embrace} the back and waist of another person”} and \textit{“One person \textbf{supports} the lower part of the other person's left \textbf{arm} with both hands”}.
These cases require fine-grained interpersonal coordination, accurate contact modeling, and temporally coherent reaction timing.

As shown in Fig.~\ref{fig:hhi} and quantitatively validated in Tab.~\ref{tab:hhi}, ReMoGen consistently produces more coordinated, semantically accurate, and physically plausible reaction motions than competing methods. Baselines often exhibit unnatural timing, misaligned contact, or unstable body dynamics, whereas ReMoGen preserves interpersonal coordination and better matches ground-truth behavior.

\begin{figure*}[t]
    \centering
    \includegraphics[width=1.0\linewidth]{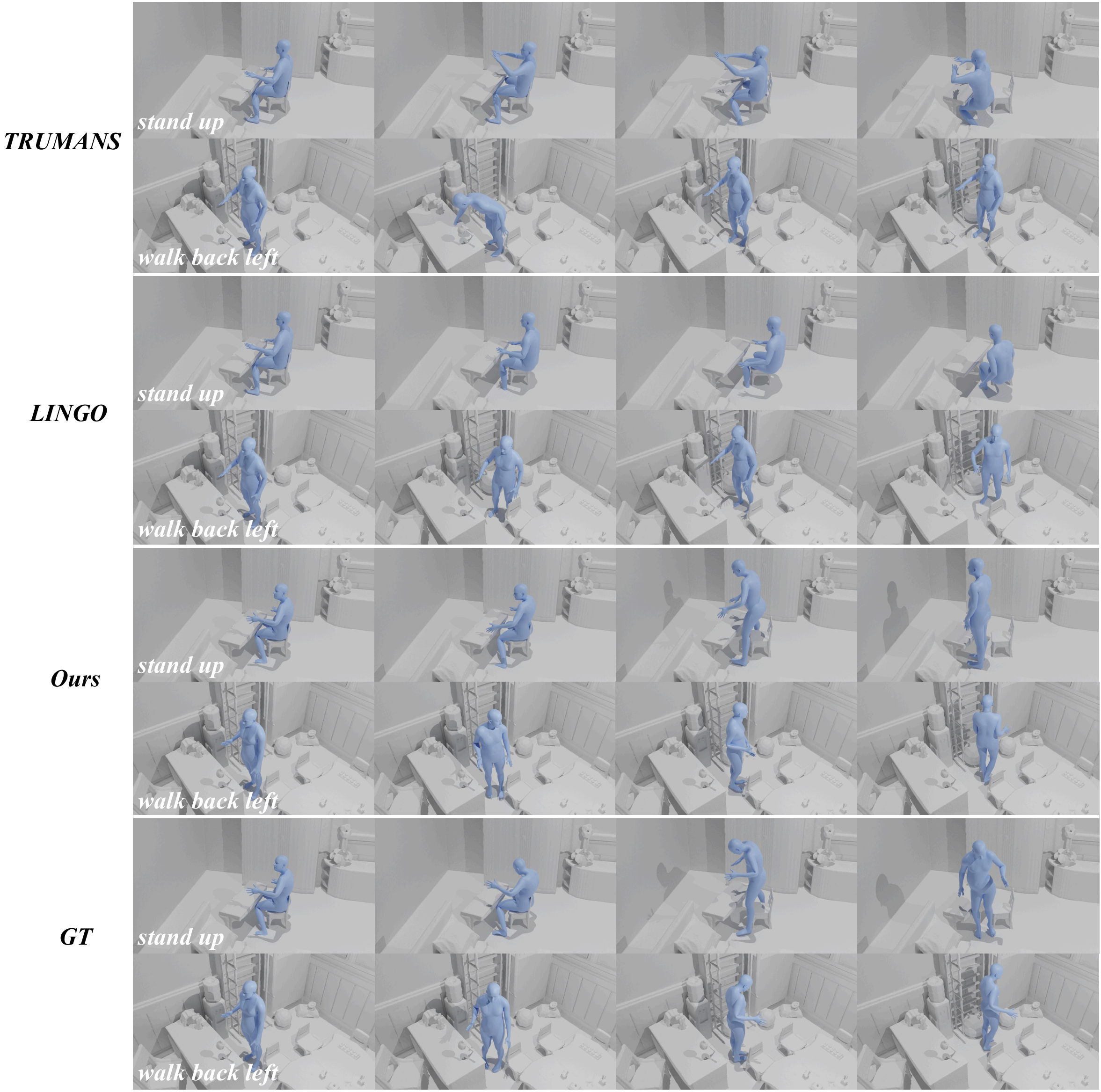}    \caption{Qualitative comparisons on Human-Scene Interaction tasks. All methods are evaluated with goal location provided.}
    \label{fig:hsi}
\end{figure*}

\paragraph{Human-Scene Interaction.}
Fig.~\ref{fig:hsi}, together with the quantitative scores in Tab.~\ref{tab:hsi}, demonstrates ReMoGen's strong scene awareness. 

We present two representative examples: \textit{“\textbf{stand up} from seat”} and \textit{“\textbf{walk back left} while holding small\_plant in right hand”}. Our model produces spatially consistent behaviors—such as standing up, walking with objects, or navigating around obstacles—while preserving motion coherence. Baselines struggle with scene grounding and often generate misaligned or unstable movements.

\subsection{Human-Human-Scene Results}
\paragraph{Zero-shot Results.}
\begin{figure*}[t]
    \centering
    \includegraphics[width=0.8\linewidth]{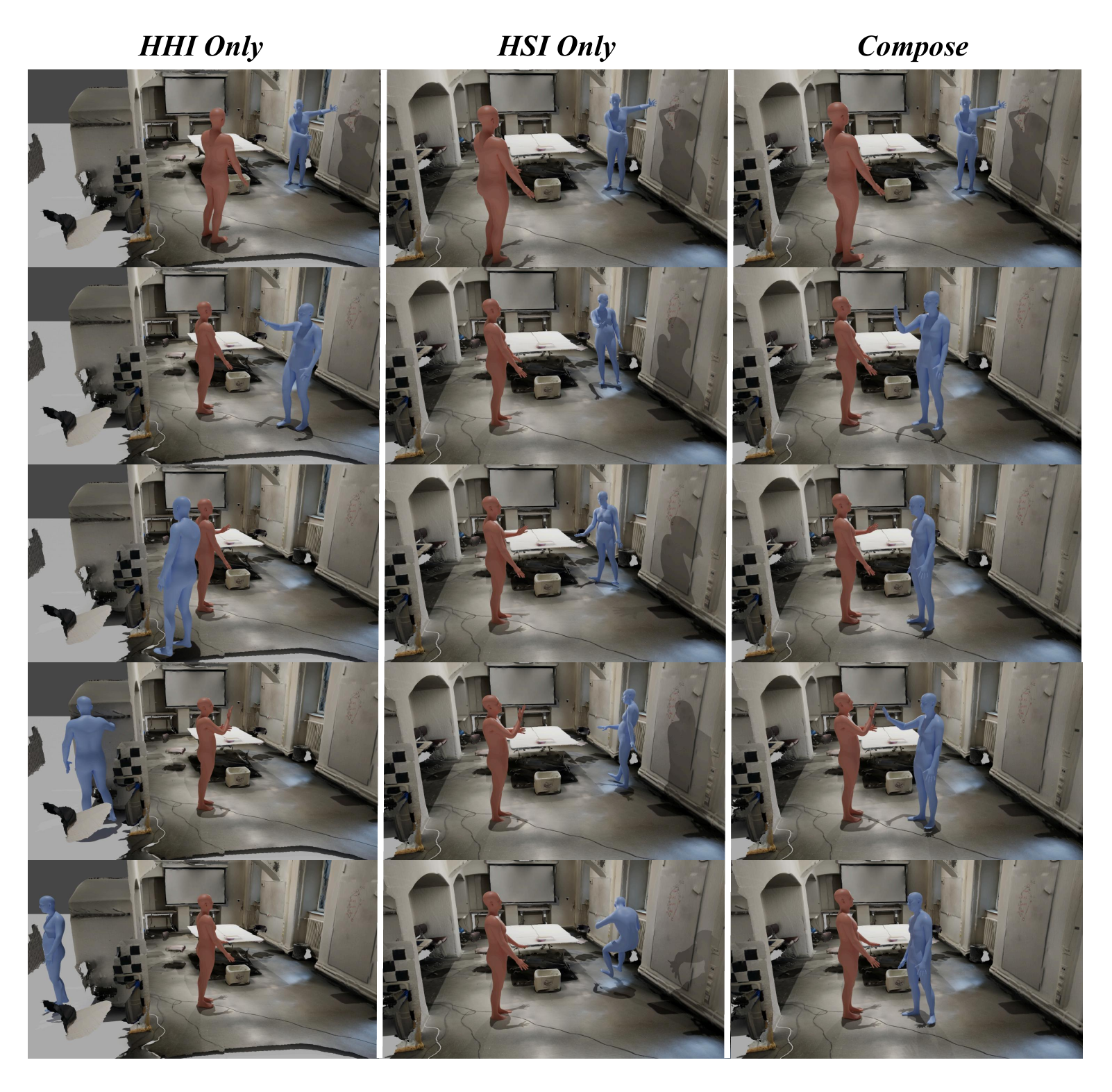}    \caption{Zero-shot Human–Human–Scene Interaction results.}
    \label{fig:hhsi_zs}
\end{figure*}
 Fig.~\ref{fig:hhsi_zs} demonstrates the zero-shot results. We present an example of the textual input \textit{“The person explains on the blackboard"}. Due to the significant domain gap, all three zero-shot results show limited semantic alignment. Using only the HHI branch or only the HSI branch leads to incomplete behavior. However, the composition of interaction modules generates coherent in-scene interactions.
\paragraph{Few-step Adaptation Results.}
\begin{figure*}
    \centering
    \includegraphics[width=1.0\linewidth]{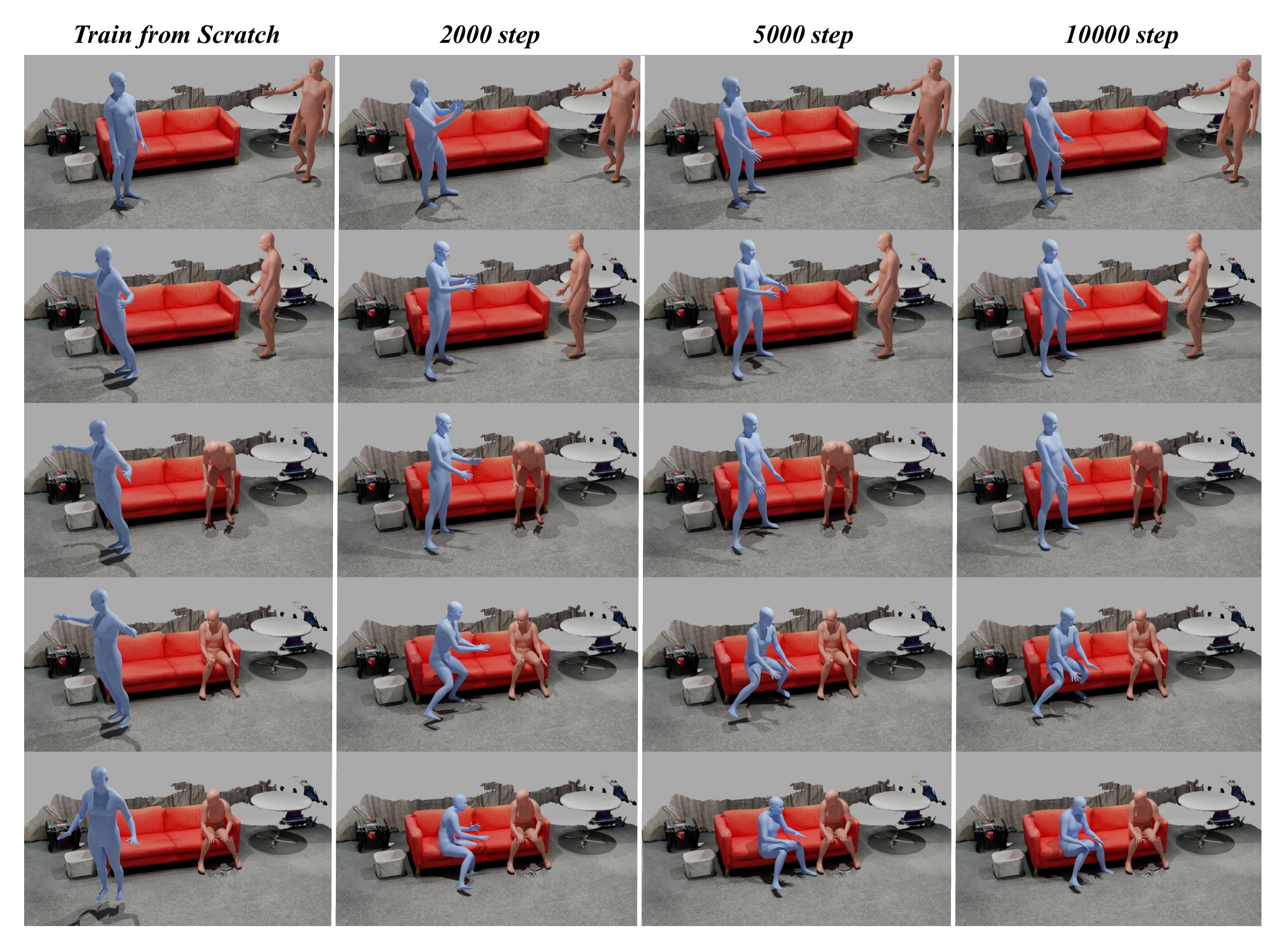}    \caption{Few-step finetuning on EgoBody.
Initializing from our universal prior enables rapid adaptation, producing natural reactions after only a small number of updates.}
    \label{fig:hhsi_ft}
\end{figure*}
We present an example: \textit{“Two people are engaged in a casual chat.”} This example includes motion implicitly involving the action of sitting down.
This scenario requires the model to recognize mixed human-human-scene cues, maintain conversational posture alignment, and correctly represent the subtle transition into a seated position.

As shown in Fig.~\ref{fig:hhsi_ft} and quantitatively verified in Tab.~\ref{tab:egobody}, ReMoGen rapidly adapts to the EgoBody dataset when initialized from the universal prior. With only 2k-10k finetuning steps, our model surpasses scratch-trained baselines that require hundreds of thousands of updates. The qualitative results illustrate that the adapted model produces stable, well-coordinated mixed-domain interactions.
\paragraph{Results in Diverse Scene.}
\begin{figure*}
    \centering
    \includegraphics[width=1.0\linewidth]{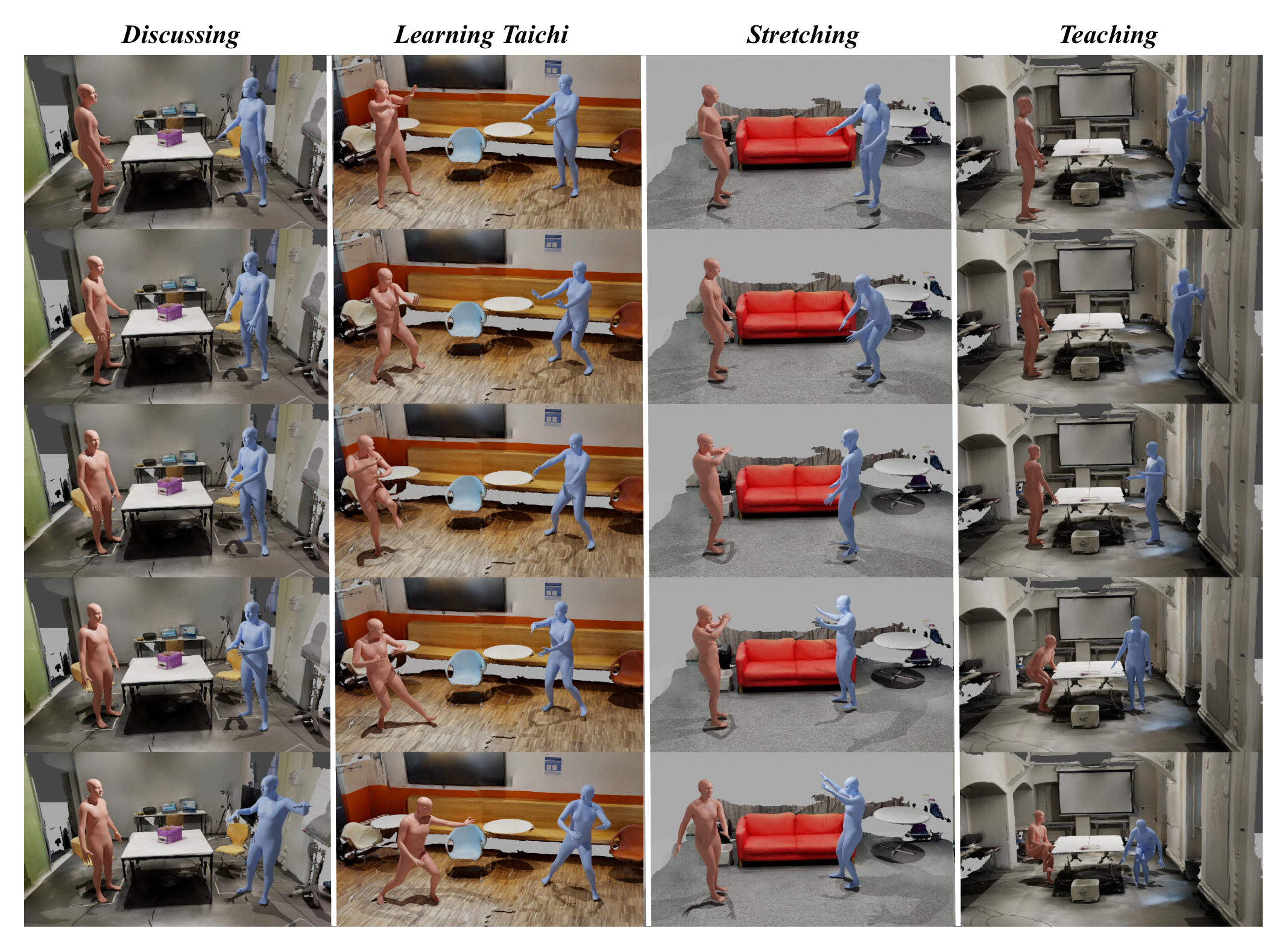}    \caption{Results in diverse Human–Human–Scene settings.
ReMoGen generates semantically rich and varied interaction behaviors across different scenes and activity types. We present the results of finetuning 65k steps.}
    \label{fig:hhsi_best}
\end{figure*}
We present four representative examples illustrating the diversity and robustness of ReMoGen across complex multi-agent and scene-rich environments: \textit{“Two individuals are engaged in a casual chat”}, \textit{“The woman is learning TaiChi”}, \textit{“A woman stretches her arms above her head”}, and \textit{“A teacher is explaining on the blackboard”}.
These scenarios span a wide range of interaction types—from individual scene-aware motions to multi-person collaborative activities.

Fig.~\ref{fig:hhsi_best} showcases ReMoGen's ability to generate diverse, semantically rich interactions in complex scenes, including Taichi movements, stretching, teaching, and conversation. These results qualitatively reflect the strong mixed-modality generalization trends reported in Tab.~\ref{tab:egobody}.

\subsection{Ablation Study Results}

\paragraph{Universal Prior Ablation.}
\begin{figure*}[t]
    \centering
    \includegraphics[width=1.0\linewidth]{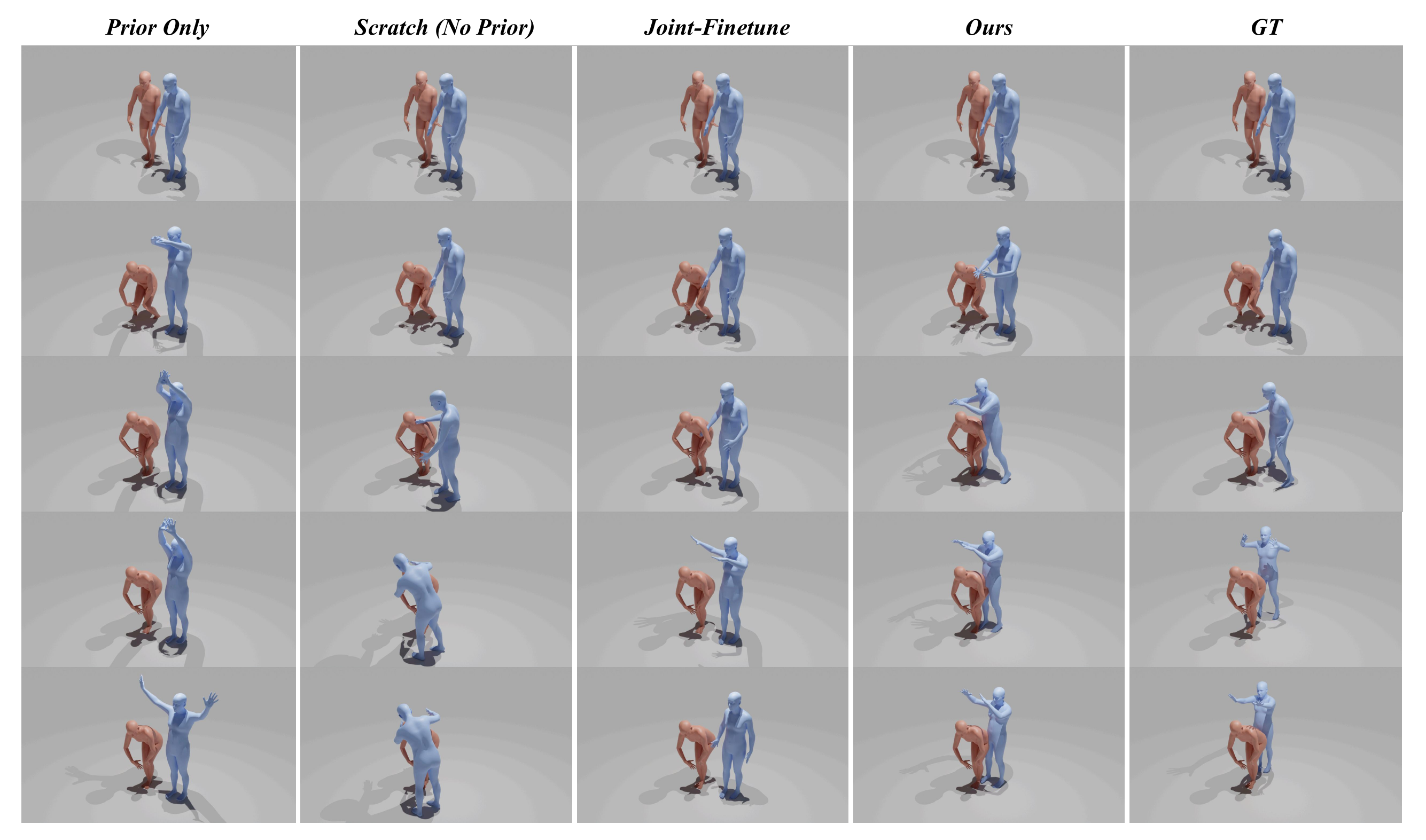}    \caption{Ablation on the Universal Motion Prior. We present \textit{"One person walks behind the other person and strikes a pose by extending both hands above the head."}
Training from scratch or joint finetuning leads to unnatural or unstable reactions, while our prior-guided modular learning produces natural, coherent motion.}
    \label{fig:abla_prior}
\end{figure*}
Fig.~\ref{fig:abla_prior} visualizes the effects observed quantitatively in Tab.~\ref{tab:abla_2stage}. Models trained from scratch suffer from limited data and exhibit low diversity and unstable dynamics. Joint finetuning erodes pretrained kinematic knowledge. Only our modular prior-guided approach preserves both natural motion structure and interaction semantics.
\paragraph{Frame-wise Segment Refinement Ablation.}
\begin{figure*}[t]
    \centering
    \includegraphics[width=1.0\linewidth]{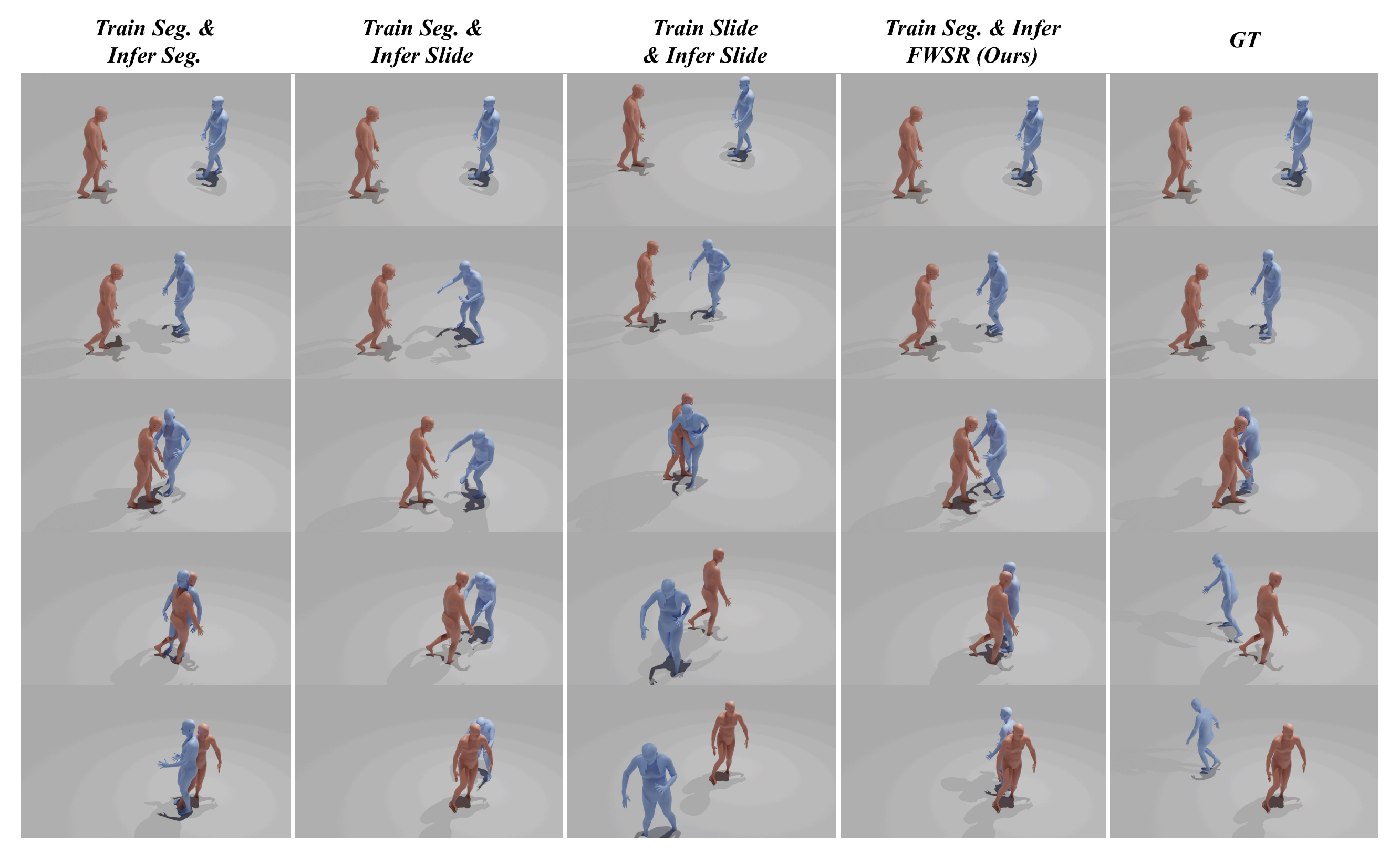}    \caption{Ablation on Frame-wise Segment Refinement. We present \textit{"A person runs towards someone and passes by them."}
FWSR provides fine-grained updates that improve responsiveness without sacrificing stability, outperforming both segment-only and naive slide-style inference.}
    \label{fig:abla_fwsr}
\end{figure*}
The qualitative comparisons in Fig.~\ref{fig:abla_fwsr} correspond directly to the latency and accuracy trends in Tab.~\ref{tab:abla_fwr}. Slide-style inference reacts quickly but introduces artifacts, while segment-only inference lacks responsiveness. Our Frame-wise Segment Refinement achieves a favorable balance—real-time responsiveness with stable, coherent motion.

\section{Limitations}
\label{sec:limitations}
While ReMoGen achieves effective performance across heterogeneous interaction domains and supports real-time reaction generation, several limitations warrant further exploration. Our framework is built upon a latent diffusion prior that abstracts motion into a compressed latent space. While effective for generalization, this design may compromise fine-grained spatial accuracy, particularly in close-contact or high-precision interactions between humans and scene objects. While our current module composition strategy is simple and intuitive, more effective training-free fusion methods may further improve performance in mixed-interaction settings.


\end{document}